\documentclass{article}
\PassOptionsToPackage{numbers, sort&compress, comma, square}{natbib}
\usepackage[preprint]{neurips_2024}

\usepackage[utf8]{inputenc} 
\usepackage[T1]{fontenc}    
\usepackage{hyperref}       
\usepackage{url}            
\usepackage{booktabs}       
\usepackage{amsfonts}       
\usepackage{nicefrac}       
\usepackage{microtype}      
\usepackage{xcolor}         
\usepackage{booktabs}
\usepackage{makecell}
\usepackage[pdftex]{graphicx}
\usepackage{amsmath}
\usepackage{booktabs} 

\usepackage{amsmath}
\usepackage{amssymb}
\usepackage{mathtools}
\usepackage{amsthm}
\usepackage{multirow}
\usepackage{adjustbox}
\usepackage{amsthm}
\usepackage[most]{tcolorbox}
\usepackage{amsmath}
\usepackage{amssymb}
\usepackage{mathtools}
\usepackage{amsthm}
\usepackage{fancyhdr}
\usepackage{subcaption}

\usepackage{caption}
\usepackage{wrapfig}
\usepackage{bm}
\usepackage[capitalize,noabbrev]{cleveref}
\usepackage{colortbl}
\usepackage{nicematrix}
\usepackage[capitalize,noabbrev]{cleveref}
\usepackage[multiple]{footmisc}
\usepackage{pifont}
\usepackage{amsthm}   
\usepackage{tcolorbox} 
\definecolor{bluex}{rgb}{0.27, 0.42, 0.81}
\definecolor{purplex}{HTML}{9564bf}
\definecolor{red3}{HTML}{C52A20}
\definecolor{red2}{HTML}{B36A6F}
\definecolor{red1}{HTML}{FFb5b5}
\definecolor{purple}{HTML}{B36A6F}
\definecolor{darkyellow}{HTML}{D5BA82}
\definecolor{blue1}{HTML}{508AB2}
\definecolor{blue2}{HTML}{C4E4E3}
\definecolor{green1}{HTML}{A1D0C7}
\definecolor{green2}{HTML}{BFF6BA}
\definecolor{green3}{HTML}{028100}
\definecolor{teal}{HTML}{508AB2}
\definecolor{purple1}{HTML}{8d3a94}

\theoremstyle{plain}

\theoremstyle{definition}

\theoremstyle{remark}

\newtcbtheorem[number within=section]{exmp}{Example}%
{colback=green2!5,colframe=blue1,fonttitle=\bfseries, left=.02in, right=.02in,bottom=.02in, top=.02in}{exmp}
\newtcbtheorem{prompt}{Example}{
colback=green2!5,colframe=blue1,fonttitle=\bfseries, left=.02in, right=.02in,bottom=.02in, top=.02in
}{prompt}
\newtcbtheorem{prmprompt}{PRM step level labeling Prompt}{
colback=green2!5,colframe=blue1,fonttitle=\bfseries, left=.02in, right=.02in,bottom=.02in, top=.02in
}{prmprompt}

\newcommand{\methodname}{\emph{Arena Learning }}

\title{
\methodname: Build Data Flywheel for LLMs Post-training via Simulated Chatbot Arena
}

\author{Haipeng Luo$^2$\thanks{\quad Equal contributions. Work done during the internship of HL at Microsoft.}  \quad Qingfeng Sun$^1$\footnotemark[1]  \quad Can Xu$^1$ \quad  Pu Zhao$^1$  \\ \\ \quad {\bf Qingwei Lin$^1$ }  \quad {\bf Jianguang Lou$^1$} \quad {\bf Shifeng Chen$^3$} \quad {\bf Yansong Tang$^2$} \quad{\bf Weizhu Chen$^1$}\\ \\
      $^1$Microsoft Corporation\\
      $^2$Tsinghua University, $^3$SIAT-UCAS
      }

\begin{document}

\maketitle

\begin{abstract}

Assessing the effectiveness of large language models (LLMs) presents substantial challenges. The method of conducting human-annotated battles in an online Chatbot Arena is a highly effective evaluative technique. However, this approach is limited by the costs and time required for human annotation. In this paper, we introduce \emph{Arena Learning}, an innovative offline strategy designed to simulate these arena battles using AI-driven annotations to evaluate battle outcomes, thus facilitating the continuous improvement of the target model through both supervised fine-tuning and reinforcement learning. \emph{Arena Learning} comprises two key elements. First, it ensures precise evaluations and maintains consistency between offline simulations and online competitions via WizardArena, a pipeline developed to accurately predict the Elo rankings of various models using a meticulously designed offline test set. Our results demonstrate that WizardArena’s predictions closely align with those from the online Arena. Second, it involves the continuous improvement of training data based on the battle results and the refined model. We establish a data flywheel to iteratively update the training data by highlighting the weaknesses of the target model based on its battle results, enabling it to learn from the strengths of multiple different models. We apply \emph{Arena Learning} to train our target model, WizardLM-$\beta$, and demonstrate significant performance enhancements across various metrics. This fully automated training and evaluation pipeline sets the stage for continuous advancements in various LLMs via post-training. Notably, \emph{Arena Learning} plays a pivotal role in the success of WizardLM-2\footnote{\quad \url{https://github.com/nlpxucan/WizardLM}}, and this paper serves both as an exploration of its efficacy and a foundational study for future discussions related to WizardLM-2 and its derivatives.

\end{abstract}

\section{Introduction}

In recent years, the field of natural language processing (NLP) has witnessed a remarkable transformation, driven by the rapid advancements in large language models (LLMs). These models, trained on vast amounts of text data, have demonstrated an exceptional ability to understand, generate, and interact with human language in a wide range of tasks ~\cite{brown2020language,ouyang2022training,touvron2023llama}. One of the most exciting applications of LLMs has been in the realm of conversational AI \cite{openai2023gpt4, anthropic2024claude, team2023gemini, Bai2023QwenTR, Deepseek}, where they have been utilized to create powerful chatbots capable of engaging in naturalistic dialogues. One of the key factors contributing to the success of LLM-powered chatbots is the ability to leverage large-scale high-quality instruction following data for  effective  post-training \cite{vicuna2023, alpaca, xu2023wizardlm, wang2023openchat, ding2023_ultrachat}. By exposing these models to a diverse range of conversational tasks and instructional scenarios, researchers have been able to imbue them with a deep understanding of how to effectively communicate and assist humans. 

\begin{figure}[!t]
\vspace{-12.5pt}
\centering
     \includegraphics[width=1\textwidth, scale=0.9, trim=0 0 2 251,clip]{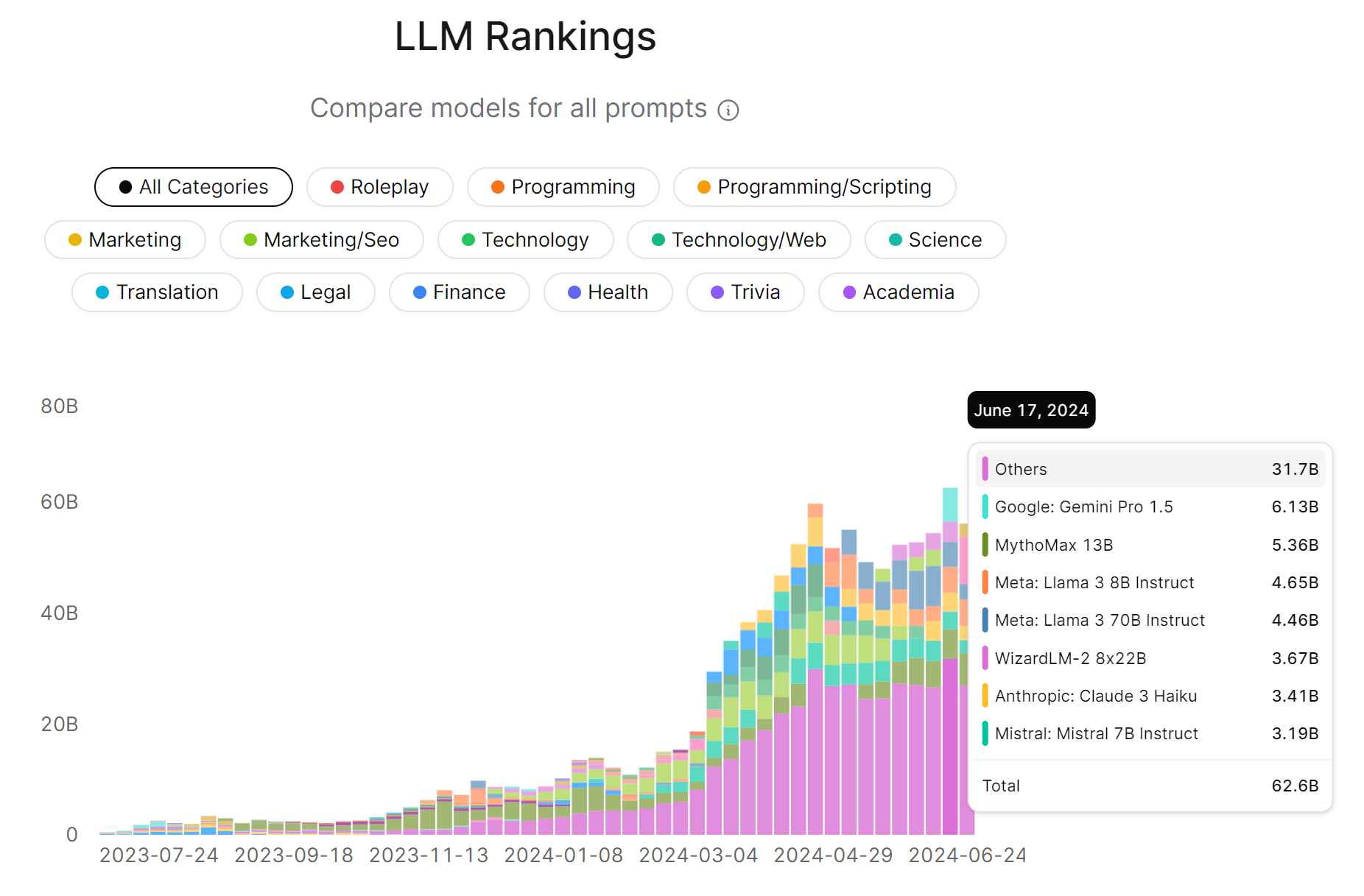}
     \caption{OpenRouter LLM Rankings on processed tokens (https://openrouter.ai/rankings).}
     \label{fig:openrouter_main_2}
     \vspace{-6pt}
\end{figure}

With the rapid implementation of various large model applications and the reduction of inference costs, the interest and demand from businesses and consumers in using large language model services have increased rapidly. As shown in the Figure \ref{fig:openrouter_main_2}, just the OpenRouter platform will process more than 60B tokens every day. At the same time, with the innovation and deepening of application scenarios, this requires those models to continue to evolve to adapt to the user's new intentions and instructions. Therefore, building an efficient data flywheel to continuously collect feedback and improve model capabilities has become a key  direction for next generation AI research.

In this context, the emergence of the LMSYS Chatbot Arena \cite{zheng2024judging_mtbench, chiang2024chatbot} has been a significant development. This is a platform that facilitates the assessment and comparison of different chatbot models by pitting them against each other in a series of conversational challenges and rank with Elo rating system \cite{bai2022training}. By leveraging a diverse set of human evaluators, the Chatbot Arena provides a more robust and comprehensive evaluation of chatbot performance, going beyond the limitations of traditional benchmarking approaches. At the same time, it also opened up some real direct chat and battle preferences data \cite{zheng2024lmsyschat1m}, which have been proven to be valuable resources for model post-training and developmental guidance \cite{zhu2023starling}. However, the human-based evaluation process poses its own challenges: Manually orchestrating and waiting the interactions between chatbots and human evaluators can be time-consuming and resource-intensive, limiting the scale and frequency of evaluation and training data opensource cycles. On the other hand, due to their priority limitations \cite{chiang2024_lmsys_chatbot}, most models are unable to participate in arena evaluations, and the community can only obtain 10\% of the chat data at most, making it hard to directly and efficiently guide the development of the target model based on this Arena. Therefore, the need for a more efficient and scalable arena-based pipeline to chatbot post-training and evaluation has become increasingly pressing.

\begin{figure}[!]
\centering
     \includegraphics[width=1\textwidth, scale=1, trim=0 90 0 20,clip]{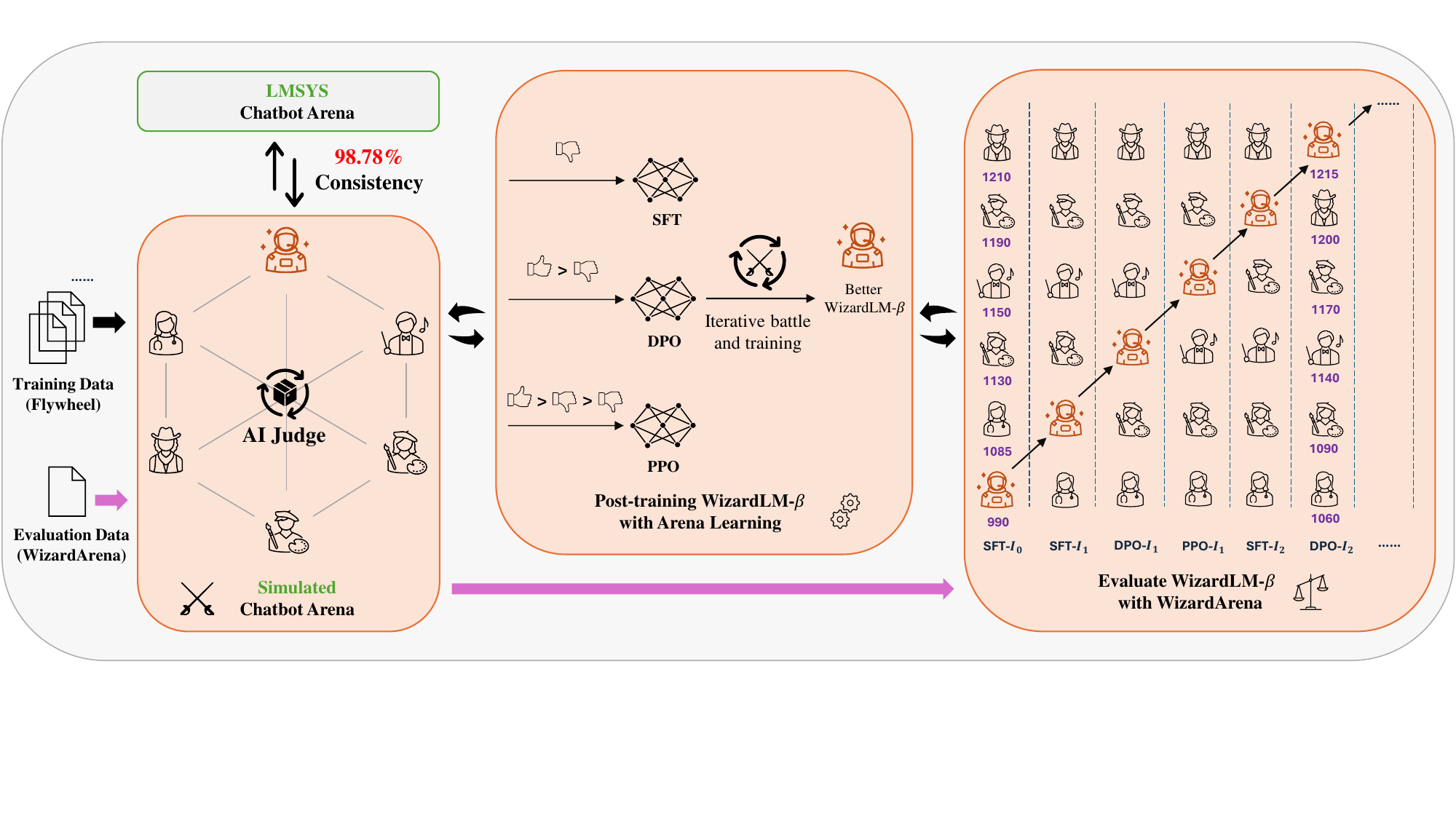}
     \caption{Overview of \methodname post-training data flywheel and WizardArena evaluation.}
     \label{fig:overview}
          \vspace{-15pt}

\end{figure}

\begin{figure}[!t]
\vspace{-15pt}
\centering
     \includegraphics[width=1\textwidth, scale=1, trim=0 100 2 40,clip]{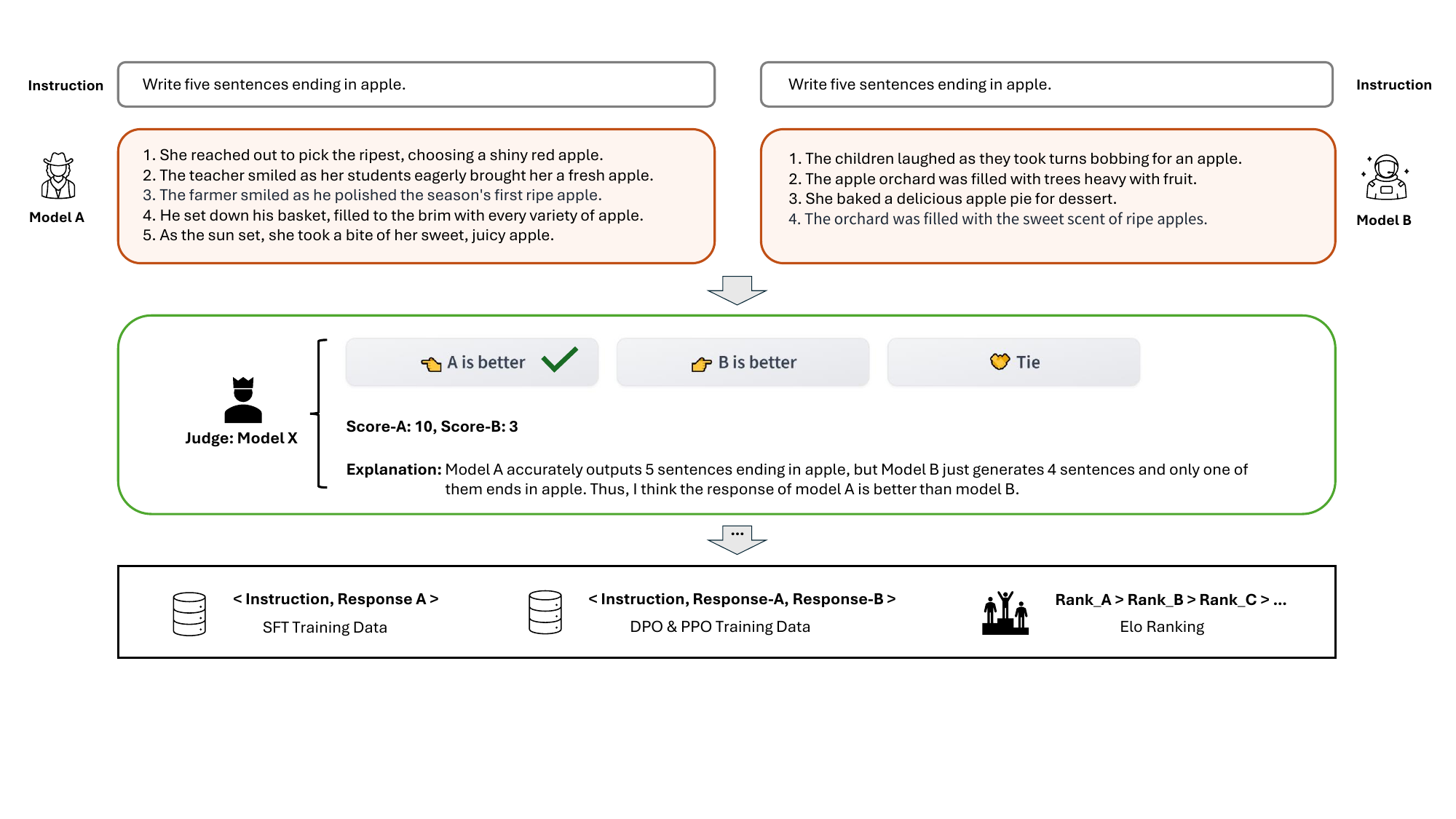}
     \caption{Overview of Running Example: how we use simulated AI-powered pair wise battle arena to produce post-training data and evaluate models.}
     \label{fig:running}
     \vspace{-15pt}

\end{figure}

To address these challenges, this paper introduces a novel approach called \emph{Arena Learning}, which is a training and evaluation pipeline fully based on and powered by AI LLMs without human evaluators. The primary objective of \methodname is to build an efficient data flywheel and mitigate the manual and temporal costs associated with post-training LLMs while retaining the benefits of arena-based evaluation and training. As the running example shown in the Figure \ref{fig:running}, the key is that \methodname simulates an offline chatbot arena, and can efficiently predict accurate performance rankings among different arena battle models based on a powerful ``judge model'', which could automatically imitate the manner of human annotators in judging a responses pair of two models and provide rankings, scores, and explanation.

In the post-training scenario, as shown in the Figure \ref{fig:overview}, \methodname simulates battles among target model (referred to as WizardLM-$\beta$) and various state-of-the-art models on a large scale of instruction data.  These synthetic battle results are then used to enhance WizardLM-$\beta$ through some training strategies, including supervised fine-tuning (SFT), direct preference optimization (DPO)  \cite{Rafailov2023_DPO}, and proximal policy optimization (PPO) \cite{ouyang2022training-2}, enabling it to learn from the strengths of other good models.  Furthermore, \methodname introduces an iterative battle and training process, where the WizardLM-$\beta$ is continuously updated and re-evaluated against SOTA models. This allows for the WizardLM-$\beta$ to iteratively improve and adapt to the evolving landscape of the arena, ensuring that it remains competitive and up-to-date with the latest top-tier competitors in the field.

In the evaluation scenario, we firstly contribute a carefully prepared offline testset - WizardArena, it effectively balances the diversity and complexity of evaluation. By automating the pair judgement process with ``judge model'', WizardArena significantly reducing the associated costs and priority limitations, and could produce the Elo rankings and detailed win/loss/tie statistics. 

The experimental results demonstrate that the Elo rankings produced by WizardArena achieve an average consistency of $98.79\%$ with the LMSys Chatbot Arena, outperforming Arena-Hard-v1.0 by $8.58\%$ and MT-Bench by $35.23\%$. This finding not only validates the effectiveness of WizardArena as a reliable and cost-effective alternative to human-based evaluation platforms, but also further proves the reliability of using the ``judge'' model to generate a large amount of battle training data in simulated arena. Moreover, the models trained on the extensive battle data generated by \methodname exhibit significant performance improvements during the SFT, DPO, and PPO stages. In three iterative loops, our model can achieve significant improvements in each round compared to the previous one, revealing that \methodname can scale up to more training data. These results highlight the value and power of \methodname in post-training, which leverages the collective knowledge and capabilities of multiple models to drive the WizardLM-$\beta$'s performance to a new height. Our main contributions are as follows:

\begin{itemize}
    \item We introduce \emph{Arena Learning}, a novel AI powered method which help us build an efficient data flywheel for large language models post-training by simulating offline chatbot arena, which leverages AI annotator to mitigate the manual and temporal costs.
    \item We contribute a carefully prepared offline testset - WizardArena, and demonstrate its high alignment with the online Elo rankings among different LLMs from human-based LMSys Chatbot Arena. 
    \item Experimental results demonstrate the effectiveness of \methodname in producing large-scale synthetic data flywheel to continuously improve WizardLM-$\beta$, through various training strategies including SFT, DPO, and PPO.
\end{itemize}


\section{Approach}

In this section, we elaborate on the details of the proposed \emph{Arena Learning}. As illustrated in Figure~\ref{fig:overview}, the closed loop pipeline mainly contains three components: Offline Pair-wise LLM Battle Arena, Iterative Post-training, and Model Evaluation.

\subsection{ChatBot Arena and Elo Ranking}

The Chatbot Arena is a pioneering platform that has revolutionized the way chatbot models are evaluated and compared.  It facilitates the assessment of different chatbot models by pitting them against each other in a series of conversational challenges. At the core of this Arena lies the concept of Elo rankings, a widely adopted rating system originally devised for chess players. Elo rankings \cite{bai2022training} are used to quantify the relative performance of chatbot models based on a series of head-to-head battles. Each model is initially assigned an arbitrary Elo rating, which is then updated after every battle based on the outcome (win, loss, or tie) and the rating difference between the competing models. If a higher-rated model defeats a lower-rated one, its Elo rating increases slightly, while the loser's rating decreases by a corresponding amount.

\subsection{Using a Powerful LLM as Judge to Simulate Human Annotators}

At the core of the simulated arena battles in \methodname lies a powerful LLM that serves as the `judge model''. This judge model is specifically prompted and adjusted by us on a diverse range of conversational pair data, enabling it to evaluate the quality, relevance, and appropriateness of the models' responses objectively and consistently. The judge model's role is to analyze and compare the responses provided by the pair battle models for each conversational sample. Specifically, to assess the response quality of each LLM, we use prompt engineering with the Llama3-70B-Chat model~\cite{touvron2023llama2}.  The inputs are dialogue history, user instruction, and the responses of two LLMs. The outputs consist of scores for each LLM, along with explanations focused on various factors, such as coherence, factual accuracy, context-awareness, and overall quality, to determine whether one response is superior to the other.  To mitigate potential position bias~\cite{zheng2024judging_mtbench, Wang2023LargeLM, arenahard2024_lmsysarena_hard}, we employ a two-game setup, alternating the positions of the two LLMs. Each model receives an overall score on a scale of 1 to 10, where a higher score reflects superior overall performance. Following, we will use this ``judge'' model in both \methodname post-training and WizardArena evaluation stages.

\subsection{Build a Data Flywheel to Post-train LLMs}

\subsubsection{Collect Large-Scale Instruction Data}

To facilitate leveraging the simulated arena battles among models to train WizardLM-$\beta$,  \methodname relies on a large-scale corpus of conversational data $D$. The data collection process involves several stages of filtering, cleaning, and deduplication to ensure the quality and diversity of the instruction data. The simulated arena battle outcomes are then used to generate training data for the WizardLM-$\beta$, tailored to different training strategies: supervised fine-tuning (SFT), direct preference optimization (DPO), and proximal policy optimization (PPO). We split the data equally into some parts $D = \{D_0, D_1, D_2, ..., D_N\}$ for following iterative training and updates respectively.

\subsubsection{Iterative Battle and Model Evolving}

 \methodname employs an iterative process for training and improving the WizardLM-$\beta$. After each round of simulated arena battles and training data generation, the WizardLM-$\beta$ is updated using the appropriate training strategies (SFT, DPO, and/or PPO). This updated model is then re-introduced into the arena, where it battles against the other SOTA models once again. This iterative process allows the WizardLM-$\beta$ to continuously improve and adapt to the evolving landscape of the arena. As the model becomes stronger, the simulated battles become more challenging, forcing the WizardLM-$\beta$ to push its boundaries and learn from the latest strategies and capabilities exhibited by the other models. Additionally, the iterative nature of  \methodname enables the researchers to monitor the progress and performance of the WizardLM-$\beta$ over time, providing valuable insights into the effectiveness of the different training strategies and potential areas for further improvement or refinement.

The following is the first training iteration $I_1$: Before that, we first train the initial version of WizardLM-$\beta$-SFT-$I_0$ with $D_0$, then select some other state-of-the-art LLMs $M$ which ranking top on WizardArena testset, following we let  WizardLM-$\beta$-SFT-$I_0$ as the competitor model, and battle with $M$ on $D_1$ , and focus on extracting instances where the WizardLM-$\beta$'s response is considered inferior to the winning model's response, as determined by the judge model. These instances are collected, and the winning model's response is used as the target output for fine-tuning the next WizardLM-$\beta$-SFT-$I_1$ model. For DPO, we use WizardLM-$\beta$-SFT-$I_1$ as competitor to battle with $M$ on $D_2$, and then we treat win and loss responses as the < choice, reject > pairs to training the WizardLM-$\beta$-DPO-$I_1$. For PPO, we leverage the same battle process between WizardLM-$\beta$-DPO-$I_1$ and $M$ on $D_3$ to obtain the < choice, reject > pairs to train the reward model and WizardLM-$\beta$-PPO-$I_1$. In the second training iteration $I_2$, we select the best WizardLM-$\beta$-PPO-$I_1$ on the WizardArena as the initial competitor model of $I_2$, and adopt similar process to train next SFT, DPO, and PPO models. Table \ref{tab:data_source_initial_model} shows the details of data and models used in each stage.

\begin{table}[ht]
\vspace{-8pt}
\centering
\caption{Data and models used in different training stages}
\scalebox{0.86}{
\begin{tabular}{@{}c|cccc@{}}
\toprule
New Model          & Train From         & Competitor Model    & Training Data                       \\ \midrule
SFT-\(I_0\)     & Mistral-Base       & -               & \(D_0\)                             \\ \midrule
SFT-\(I_1\)      & Mistral-Base           & SFT-\(I_0\)     & \(D_0 \cup D_1\)                    \\
DPO-\(I_1\)     & SFT-\(I_1\)           & SFT-\(I_1\)     & \(D_2\)                             \\
PPO-\(I_1\)    & DPO-\(I_1\)           & DPO-\(I_1\)     & \(D_3\)                             \\ \midrule
SFT-\(I_2\)     & Mistral-Base       & PPO-\(I_1\)     & \(D_0 \cup D_1 \cup D_4\)           \\
DPO-\(I_2\)     & SFT-\(I_2\)           & SFT-\(I_2\)     & \(D_2 \cup D_5\)                    \\
PPO-\(I_2\)    & DPO-\(I_2\)           & DPO-\(I_2\)     & \(D_3 \cup D_6\)                    \\ \midrule
SFT-\(I_3\)    & Mistral-Base       & PPO-\(I_2\)     & \(D_0 \cup D_1 \cup D_4 \cup D_7\)  \\
DPO-\(I_3\)    & SFT-\(I_3\)           & SFT-\(I_3\)     & \(D_2 \cup D_5 \cup D_8\)           \\
PPO-\(I_3\)     & DPO-\(I_3\)           & DPO-\(I_3\)     & \(D_3 \cup D_6 \cup D_9\)           \\ \bottomrule
\end{tabular}}
\label{tab:data_source_initial_model}
\vspace{-8pt}
\end{table}

\subsection{Evaluate LLMs with WizardArena}


To accurately evaluate the performance of chatbot models and predict their Elo rankings, \methodname relies on a carefully curated offline test set, which is designed to strike a balance between diversity and complexity~\cite{zheng2024judging_mtbench, arenahard2024_lmsysarena_hard, alpaca_eval}, ensuring a comprehensive assessment of the models' capabilities across a wide range of conversational scenarios. Inspired by WizardLM~\cite{xu2023wizardlm} In-Breadth Evolving and In-Depth Evolving, we construct the following two subsets:

\noindent\textbf{Diverse Subset} The diverse subset of the test set is constructed to capture a broad range of topics, styles, and conversational contexts. To achieve this, we employs text clustering techniques on a large corpus of instructions and conversational data. The clustering process begins by representing all the instructions in a conversation as a high-dimensional vector using state-of-the-art embedding models (i.e., gte-large~\cite{Li2023TowardsGT}). These vectors capture the semantic and contextual information within the text, enabling the clustering algorithm to group similar samples together. Once the clustering is complete, we selects a representative sample from each cluster, ensuring that the diverse subset of the test set capture a broad range of scenarios. This approach helps to mitigate potential biases or blindspots that may arise from relying solely on simply random sampling.

\noindent\textbf{Hard Subset} This subset is specifically designed to challenge the capabilities of even the most advanced chatbot models. To construct this subset, we leverages the power of LLMs to predict the difficulty level of each instruction. We then selects the top-ranking samples according to the predicted difficulty scores, ensuring that the hard subset of the test set comprises the most challenging and complex scenarios. This data serves as a rigorous benchmark for evaluating the robustness and capability of chatbot models in handling intricate and nuanced conversational tasks.

With the above ``judge'' model and the offline WizardArena test set in place, we proceeds to evaluate the performance of various chatbot models through a series of pair-wise battles. The outcomes of the battles are then used to compute the Elo rankings of the participating chatbot models. WizardArena adopts the same Elo rating system used in LMSYS Chatbot Arena, which has proven effective in ranking players or entities based on their head-to-head performance.

\section{Experiments}

\subsection{Experimental Setup}

\textbf{Training Data.}  We random sample 10k ShareGPT data to train a initial model WizardLM-$\beta$-$I_0$. We then collected some instructions from open available datasets \cite{alpaca, longpre2023flan, zheng2024lmsyschat1m,  xu2023wizardlm, OpenOrca},
and optimized them using the following steps: first, we filtered out all illegal and toxic conversations; second, we removed conversations with instruction lengths of less than 10; third, we eliminated duplicate instructions with prefixes of 10; next, we employed the MinHashLSH technique~\cite{Shrivastava2014-minhash} for data deduplication; subsequently, we used an embedding model gte-large ~\cite{Li2023TowardsGT} to exclude instructions from the top 5 matches in semantic similarity with benchmarks (i.e.,  WizardArena, Arena-Hard Auto~\cite{arenahard2024_lmsysarena_hard}, MT-Bench~\cite{zheng2024judging_mtbench}, AlpacaEval~\cite{alpaca_eval}, OpenLLM Leaderboard~\cite{open-llm-leaderboard, allenai:arc, hendrycks2020measuring, zellers2019hellaswag, lin2021truthfulqa}) to prevent test data leakage. Finally, we removed all non-English instructions. After completing these steps, we obtain the refined 276K dataset $D$, and randomly split it to 9 parts.

\begin{figure}[ht]
\vspace{-12.5pt}
    \centering
    \begin{minipage}{0.45\textwidth}
        \centering
        \includegraphics[width=1\linewidth]{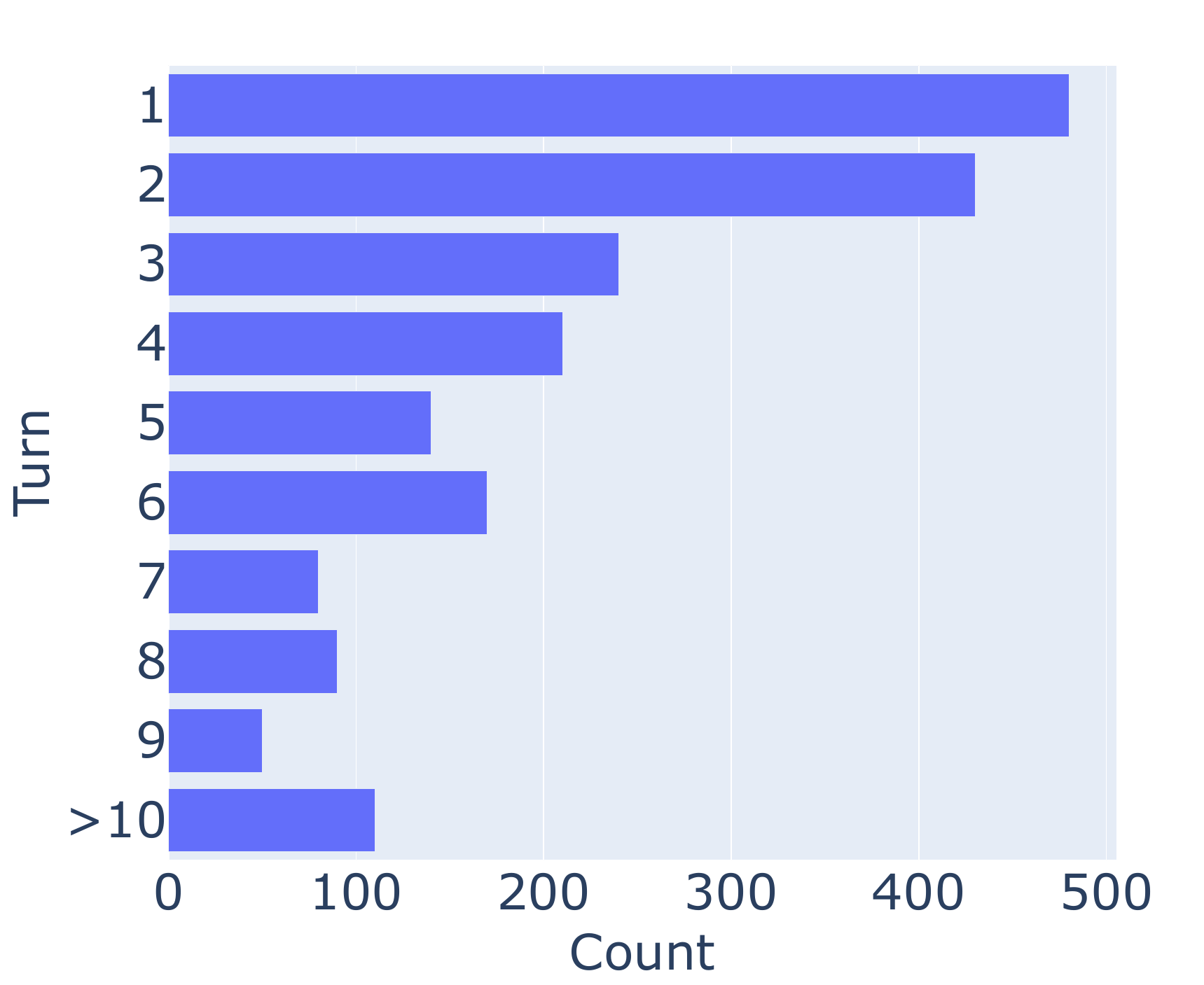}
        \caption{WizardArena-Mix Turn statistics}
        \label{fig:turn_statistics}
    \end{minipage}\hfill
    \begin{minipage}{0.55\textwidth}
        \centering
        \includegraphics[width=1\linewidth]{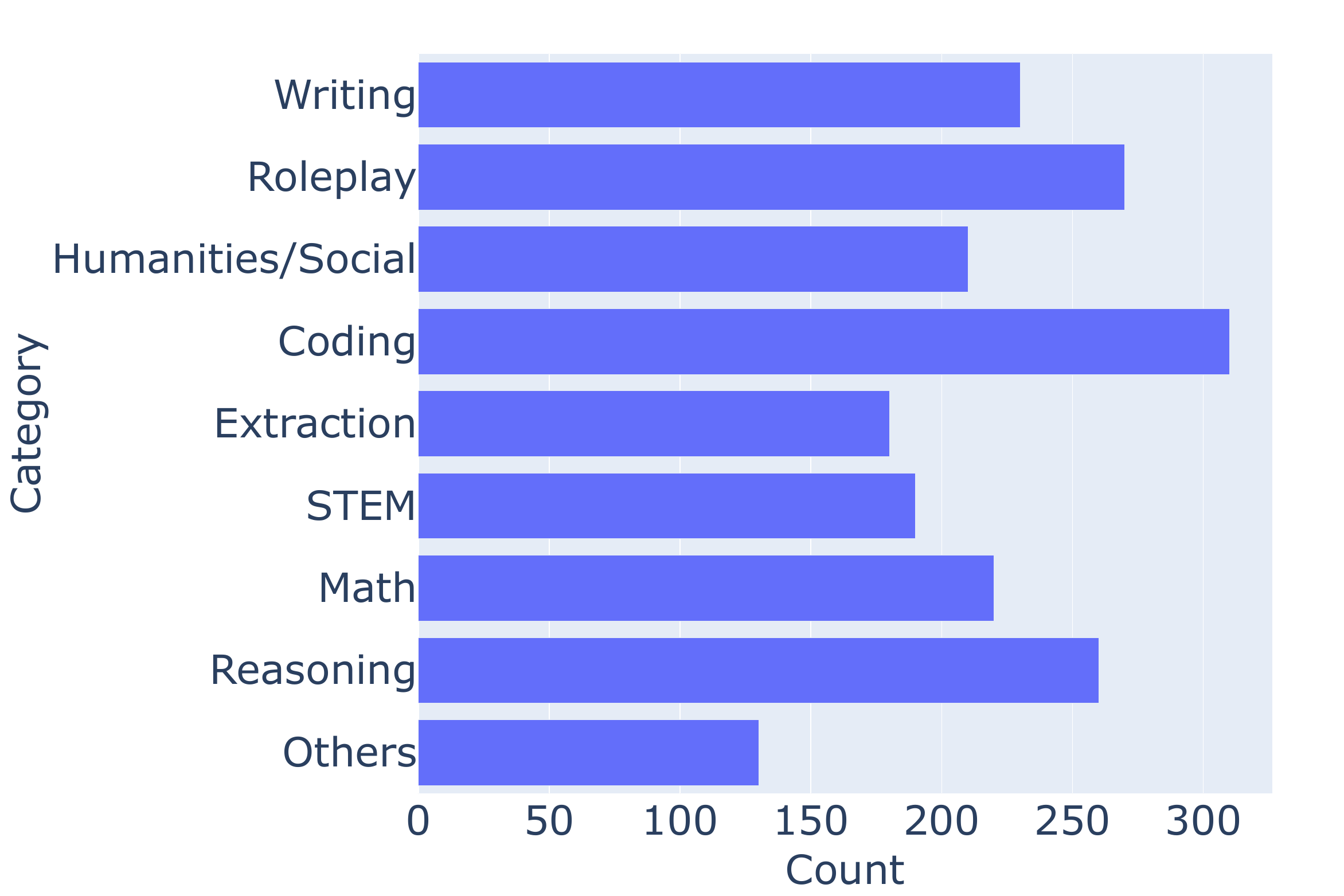}
        \caption{WizardArena-Mix Category statistics}
        \label{fig:category_testset_statistics}
    \end{minipage}
\end{figure}

\textbf{Offline Diverse \& Hard WizardArena test set.} Firstly, we processed the source data using K-Means clustering into 500 categories. From each category, we randomly selected two samples to construct 1,000 diversity samples, named as the Offline-Diverse WizardArena. Additionally, 20 samples from each category were selected at random to form a data set of 10,000 entries, we then used GPT-4-1106-preview to rate each instruction on a difficulty scale from 0 to 10 in descending order, and selected the top 1,000 entries to create the hard test set, denoted as the Offline-Hard WizardArena. The Offline-Mix WizardArena combines the Diverse and Hard test sets in 2,000 samples. Different from Arena-Hard-v1.0~\cite{arenahard2024_lmsysarena_hard}, which mainly focuses on single-turn dialogue data, WizardArena-Mix incorporates multi-turn dialogue data. Figures \ref{fig:turn_statistics} and \ref{fig:category_testset_statistics} display the distribution of dialogue turn and the categories statistics within WizardArena-Mix, respectively. The data indicates that our multi turn conversation data accounts for a large proportion, and the distribution of topics is also diverse.

\begin{wraptable}{r}{0.5\textwidth}
\vspace{-12.5pt}
    \centering
    \caption{Efficiency Comparison of LMSYS ChatBot Arena and WizardArena.}
    \scalebox{0.88}{
    \setlength{\tabcolsep}{1pt}

    \begin{tabular}{l|cc}
        \toprule
        Metrics &LMSYS ChatBot Arena & Ours \\
        \midrule
        Battle Method & Human & LLM \\
        Battle Count & 1M & 1M \\
        GPU Count & - & 16 \\
        Inference Time & - & 3 Days \\
        Judge Time & $\sim$1 year & 6 Days \\ \midrule
        Speed Up & 1x & 40x\\
        \bottomrule
    \end{tabular}
    \label{tab:comparison_total}
}
\end{wraptable}

\textbf{LLM Battle.} We selected some popular models and conducted pairwise battles in the Offline-Mix WizardArena. Llama3-70B-Instruct~\cite{touvron2023llama2} served as the ``judge'' model, with the higher-scoring model declared the winner.  Following LMSYS Chatbot Arena, we adopt the Bradley-Terry model~\cite{Fageot2023Bradley-Terry} to calculate the final ELO scores for each model. To mitigate potential position bias, we used a two-game setup, swapping the models between the first and second positions for each instance~\cite{Wang2023LargeLM}. We use multiple bootstraps (i.e., 100), and select the median as the model's ELO score. The 95\% CI is determined from the 2.5\% to 97.5\% range of confidence interval scores. Table ~\ref{tab:comparison_total} contrasts the differences between WizardArena and LMSYS Arena. WizardArena leverages LLM to conduct Battles, whereas LMSYS ChatBot Arena relies on human annotation. At the same battle count, if we use sixteen 80G GPUs for inference and judgement, the process will be completed in 9 days, achieving a 40x speedup increase compared to the 12 months required by LMSYS ChatBot Arena.

\textbf{Implementation Details.} We apply our method to the Mistral-7B~\cite{jiang2023mistral} and Mixtral-8x22B for post-training, using Llama3-70B-Instruct~\cite{touvron2023llama2} as judge models. For  WizardLM-$\beta$-7B, the battle models are \{Command R+~\cite{cohere}, Qwen1.5-72B-chat~\cite{Bai2023QwenTR}, OpenChat-3.5~\cite{wang2023openchat}\}, for WizardLM-$\beta$-8x22B, the battle models are \{GPT-4o~\cite{openai2023gpt4}, GPT-4-1106-preview~\cite{openai2023gpt4}, WizardLM-2-8x22B-0415~\cite{xu2023wizardlm}\}. In supervised fine-tuning, we trained three epochs with a learning rate of 5e-6, a batch size of 128, and a sequence length of 4096. For PPO reward model training, Mistral-7B was trained for one epoch at a learning rate of 1e-6. In PPO training, the learning rate was 1e-7 for one epoch with a KL coefficient of 0.4, and for DPO training, it was 5e-7 for two epochs with a beta of 0.3.

\subsection{Offline WizardArena closely align with the Online LMSYS ChatBot Arena. }

\begin{wrapfigure}{r}{0.60\textwidth}
\vspace{-20pt}
    \centering
        \includegraphics[width=0.5\textwidth]{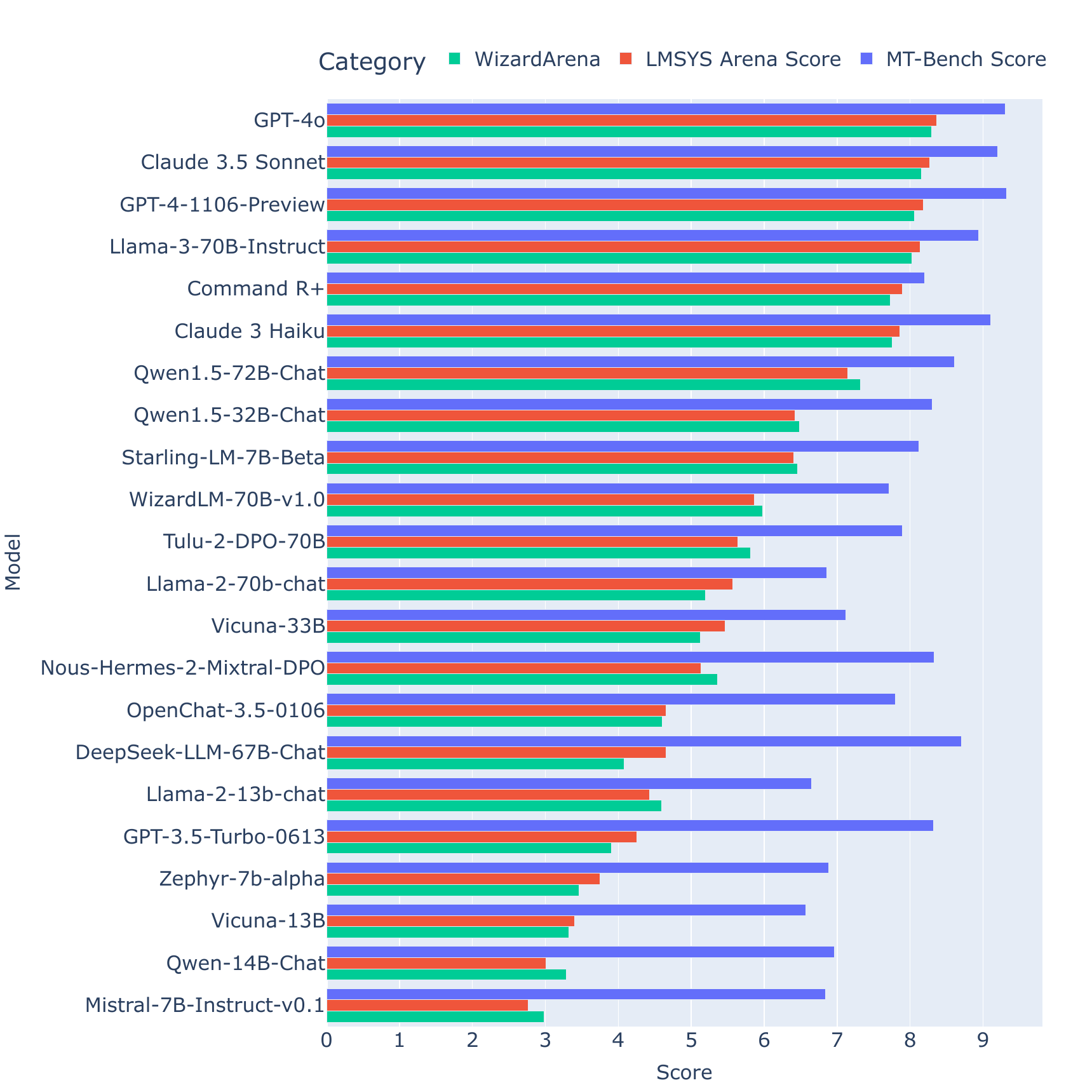}
        \caption{\footnotesize The performance of LLMs across MT-Bench, normalized LMSYS ChatBot Arena, and WizardArena.
        }
        \label{fig:arena_score_fig}
\end{wrapfigure}

Figure ~\ref{fig:arena_score_fig} and Table ~\ref{tab: arena_elo} present the rankings for some popular models across several evaluation benchmarks: LMSYS ChatBot Arena-EN~\cite{chiang2024_lmsys_chatbot}, MT-Bench~\cite{zheng2024judging_mtbench}, and WizardArena. The results reveal that employing the LMSYS ChatBot Arena as the reference benchmark in the real-world scenarios, WizardArena displays the good ranking consistency, however  MT-Bench shows the large fluctuations. In addition, there is a significant difference in performance between WizardArena diverse and hard subsets: Vicuna-33B~\cite{vicuna2023} and Qwen1.5-32B-Chat~\cite{Bai2023QwenTR} are more effective in diverse tasks, while Tulu-2-DPO-70B~\cite{ivison2023_tulu2} and Nous-Hermes-2-Mixt-DPO~\cite{Nous-Hermes-2-Mixtral-8x7B-DPO} achieves better results in hard tasks. We therefore use WizardArena-Mix as the final evaluation benchmark of \methodname to balance the strengths of different models.

\begin{table}[ht]
\vspace{-8pt}
    \centering
    \caption{The consistency of MT-Bench, Arena-Hard-v1.0, and WizardArena compared with LMSYS ChatBot Arena. Llama-3-70B-Chat is the ``Judge'' model.}
    \scalebox{0.8}{
    \setlength{\tabcolsep}{1pt}

    \begin{tabular}{l|c|c|ccc}
        \toprule
        Metrics & MT-Bench & Arena-Hard-v0.1 & \makecell{WizardArena- \\ Diverse}  & \makecell{WizardArena- \\ Hard} & \makecell{WizardArena- \\ Mix}\\
        \midrule
        Data Size & 160 & 500 & 1000 & 1000 & 2000 \\        \midrule
        Spearman Correlation & 79.36\% & 90.44\% & 98.79\% & 98.84\% & \textbf{99.23\%} \\
        Human Agreement with 95\% CI & 26.04\% & 80.86\% & 97.33\% & 98.22\% & \textbf{99.11\%} \\
        Differentiation with 95\% CI & 23.45\% & 92.33\% & 97.63\% & 96.84\% & \textbf{98.02\%} \\
        Avg.    & 42.95\% & 87.88\% & 97.92\% & 97.97\% & \textbf{98.79\%} \\
        \bottomrule
    \end{tabular}
    \label{tab:spearman_transposed}
}
\end{table}

Table ~\ref{tab:spearman_transposed} illustrates that the Offline WizardArena-Mix significantly outperforms MT-Bench across several consistent metrics which refer to the Appendix~\ref{appendix:consistency_metrics} for details:  a 19.87\% higher Spearman Correlation, a 73.07\% increase in Human Agreement with 95\% CI, and a 74.57\% improvement in Differentiation with 95\% CI. It achieves an average consistency of 98.79\% with the LMSYS ChatBot Arena by human judgment, outperforming Arena-Hard-v1.0~\cite{arenahard2024_lmsysarena_hard} by 10.91\% and MT-Bench~\cite{zheng2024judging_mtbench} by 55.84\%.  In contrast to MT-Bench and Arena-Hard-v1.0  which use proprietary models (i.e. GPT-4) as the judge model, our approach employs current SOTA open-source model Llama-3-70B-Chat, which not only has a significantly lower cost but also achieves strong consistency. Moreover, the Offline WizardArena-Mix, which integrates both Diverse and Hard test sets, achieves 0.87\% higher average consistency compared to WizardArena-Diverse and 0.82\% higher than WizardArena-Hard. This indicates that balancing diversity and complexity is crucial for the effective offline evaluation of large language models. Above results also further prove the feasibility of using the “judge” model to judge the battles between LLMs and generate a large amount of  post-training data in simulated arena.

\newpage

\begin{table}
\vspace{-18pt}

\centering
\caption{ The ELO rankings on LMSYS ChatBot Arena EN (June, 2024), MT-Bench, and WizardArena. Llama-3-70B-Chat is the ``judge''. Llama-2-70B-Chat Elo is the reference.  }
\scalebox{0.68}{
\setlength{\tabcolsep}{5pt}
\label{tab: arena_elo}

\begin{tabular}{lllllc}
\toprule
Model & \makecell{LMSYS-ChatBot \\ Arena-ELO-EN \\ (95\% CI)} & \makecell{WizardArena \\ Diverse-ELO \\ (95\% CI)} & \makecell{WizardArena \\ Hard-ELO \\ (95\% CI)} & \makecell{WizardArena\\ Mix-ELO \\ (95\% CI)} & MT-bench \\

\midrule
GPT-4o~\cite{openai2023gpt4} & 1266 (+4/-4) & 1401 (+3/-4) & 1392 (+4/-5) & 1395 (+5/-4)& 9.30 \\
Claude 3.5 Sonnet~\cite{anthropic2024claude} & 1246 (+4/-7) & 1389 (+5/-6) & 1378 (+6/-6) & 1384 (+6/-4) & 9.20 \\
Gemini 1.5 Pro ~\cite{team2023gemini} &  1235 (+5/-4) & 1383 (+6/-5) &  
 1373 (+5/-5)& 1377 (+5/-5)& - \\
GPT-4-1106-Preview ~\cite{openai2023gpt4} & 1232 (+3/-4) & 1369 (+3/-5) & 1376 (+6/-4) & 1374 (+4/-3)& 9.32 \\
WizardLM-2-8x22B-0415 ~\cite{xu2023wizardlm} & - & 1365 (+6/-7) & 1359 (+5/-7) & 1361 (+5/-6)& 9.12 \\
Llama-3-70B-Instruct ~\cite{touvron2023llama2}  & 1227 (+3/-3) & 1366 (+5/-5) & 1354 (+6/-5) & 1357 (+6/-4)& 8.94 \\
\textbf{WizardLM-$\bm{\beta}$-8x22B-$\bm{I_3}$} & - & 1355 (+5/-7) & 1346 (+6/-5) & 1349 (+5/-7)& 8.85 \\

Command R+~\cite{cohere} & 1163 (+4/-4) & 1351 (+9/-6) & 1327 (+8/-6) & 1337 (+6/-4) & 8.20 \\

Claude 3 Haiku ~\cite{anthropic2024claude} & 1158 (+4/-3) & 1340 (+4/-5) & 1345 (+5/-5) & 1342 (+4/-6) & 9.10 \\

\textbf{WizardLM-$\bm{\beta}$-8x22B-$\bm{I_2}$} & - & 1339 (+6/-6)  & 1326 (+6/-8) & 1332 (+6/-7)& 8.49 \\

Qwen1.5-72B-Chat~\cite{Bai2023QwenTR} & 1135 (+3/-4) & 1332 (+9/-7) & 1312 (+7/-5) & 1321 (+6/-5) & 8.61 \\
\textbf{WizardLM-$\bm{\beta}$-8x22B-$\bm{I_1}$} & - &  1325 (+8/-6)& 1311 (+7/-7) & 1318 (+8/-7) & 7.98 \\

Qwen1.5-32B-Chat~\cite{Bai2023QwenTR} & 1109 (+4/-5) & 1298 (+7/-8) & 1276 (+5/-8) & 1283 (+6/-4) & 8.30 \\
\textbf{WizardLM-$\bm{\beta}$-7B-$\bm{I_3}$} & - &   1269 (+5/-4) &  1278 (+5/-4) &   1274 (+5/-6) & 8.16 \\

Starling-LM-7B-Beta ~\cite{zhu2023starling} &  1108 (+5/-5)& 1275 (+6/-4) & 1270 (+6/-5) & 1272 (+4/-6) & 8.12 \\

\textbf{WizardLM-$\bm{\beta}$-7B-$\bm{I_2}$} & - & 1256 (+5/-7) & 1233 (+4/-7) & 1246 (+6/-5) & 7.98 \\
\textbf{WizardLM-$\bm{\beta}$-7B-$\bm{I_1}$} & - & 1228 (+4/-6) & 1201 (+6/-8) & 1214 (+5/-8)& 7.74 \\

WizardLM-70B-v1.0~\cite{xu2023wizardlm} & 1098 (+7/-6) & 1184 (+6/-6) & 1163 (+6/-5) & 1169 (+5/-5) & 7.71 \\
Llama-2-70B-Chat~\cite{touvron2023llama2} & 1097 (+5/-4) & 1100 (+0/-0) & 1100 (+0/-0) & 1100 (+0/-0) & 6.86 \\
Tulu-2-DPO-70B~\cite{ivison2023_tulu2} & 1091 (+8/-10) & 1147 (+8/-6) & 1181 (+5/-6) & 1157 (+4/-6) & 7.89 \\
Vicuna-33B~\cite{vicuna2023} & 1086 (+6/-5) & 1113 (+5/-7) & 1076 (+7/-5) & 1091 (+4/-5) & 7.12 \\

Nous-Hermes-2-Mixtral-DPO~\cite{Nous-Hermes-2-Mixtral-8x7B-DPO} & 1078 (+9/-8) & 1107 (+8/-6) & 1121 (+7/-7) & 1114 (+5/-4) & 8.33 \\

OpenChat-3.5~\cite{wang2023openchat} & 1065 (+9/-10) & 1042 (+7/-5) & 1050 (+8/-5) & 1045 (+5/-5) & 7.80 \\
DeepSeek-LLM-67B-Chat~\cite{Bi2024DeepSeekLS} & 1065 (+12/-10) & 991 (+7/-7) & 1008 (+5/-7) & 1000 (+7/-5) & 8.70 \\

Llama-2-13B-Chat~\cite{touvron2023llama2} & 1061 (+5/-6) & 1052 (+5/-6) & 1041 (+7/-7) & 1042 (+5/-4) & 6.65 \\
GPT-3.5-Turbo-1106~\cite{openai2023gpt4} & 1052 (+5/-5) & 955 (+6/-7) & 1004 (+6/-7) & 981 (+5/-5) & 8.32 \\
Zephyr-7b-alpha~\cite{Tunstall2023ZephyrDD} & 1040 (+17/-13) & 905 (+7/-6) & 967 (+6/-8) & 939 (+4/-5) & 6.88 \\

Vicuna-13B~\cite{vicuna2023} & 1029 (+6/-5) & 934 (+6/-7) & 923 (+8/-5) & 927 (+5/-5) & 6.57 \\

Qwen-14B-Chat~\cite{Bai2023QwenTR} & 1017 (+9/-10) & 916 (+5/-7) & 932 (+6/-8) & 924 (+4/-6) & 6.96 \\
Mistral-7B-Instruct-v0.1~\cite{jiang2023mistral} & 1009 (+7/-7) & 883 (+6/-7) & 904 (+6/-9) & 894 (+4/-5) & 6.84 \\
\textbf{WizardLM-$\bm{\beta}$-8x22B-$\bm{I_0}$} & - & 873 (+5/-9) & 897 (+4/-8) & 889 (+4/-9)& 6.78 \\
\textbf{WizardLM-$\bm{\beta}$-7B-$\bm{I_0}$} & - & 862 (+8/-7) & 884 (+6/-7) & 871 (+5/-8) & 6.41 \\
\bottomrule
\end{tabular}
}

\end{table}

\begin{figure}[!t]
\vspace{-5pt}
    \centering
       \includegraphics[width=0.85\textwidth, trim=0 210 0 0,clip]{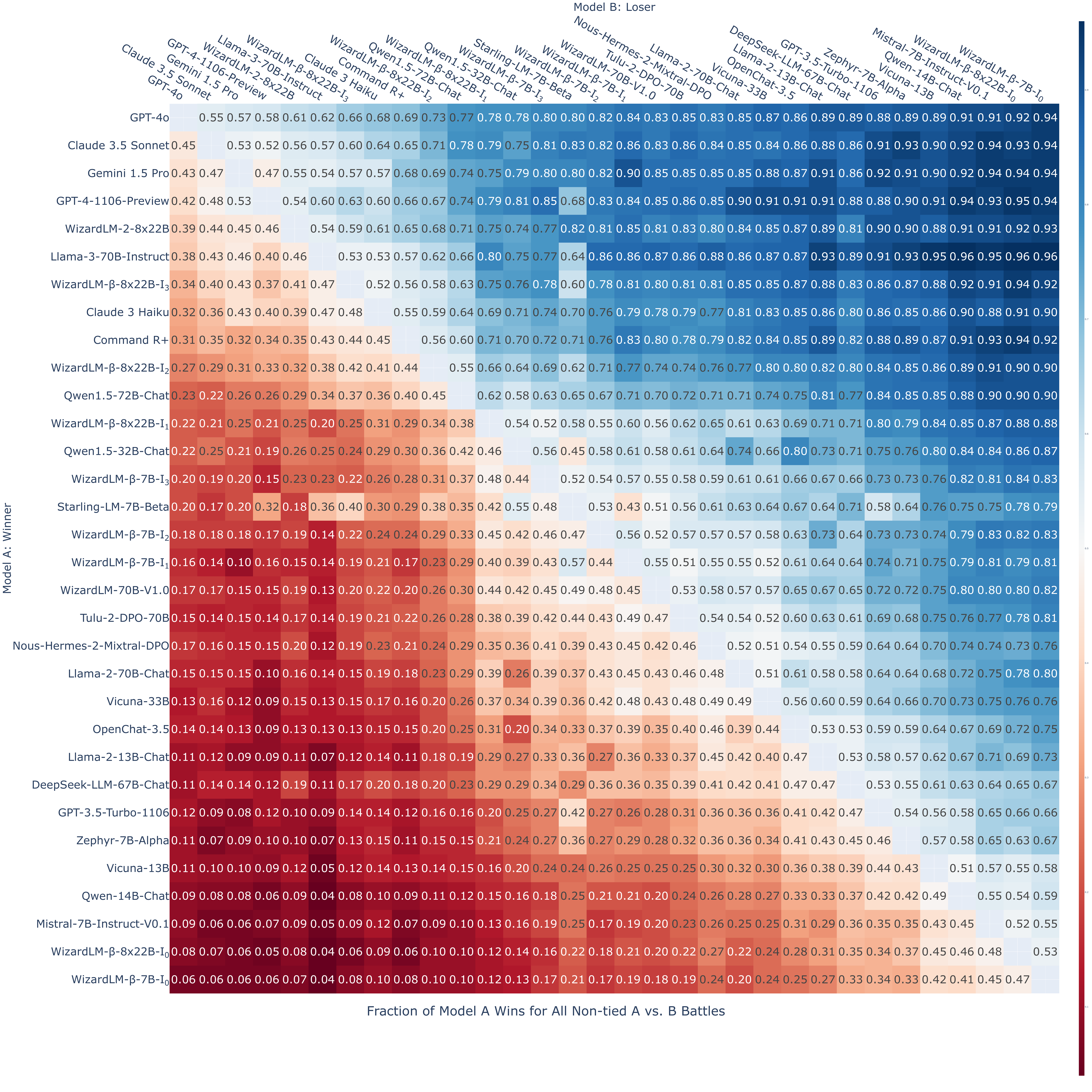}
        \caption{\footnotesize 
        Win rates (w/o tie) of models in WizardArena-Mix. Each model involved in 2k x 31 battles. }
        \label{fig:win_rate}
\end{figure}

\newpage

\begin{figure}
\vspace{-12pt}
    \centering
    \includegraphics[width=1\linewidth]{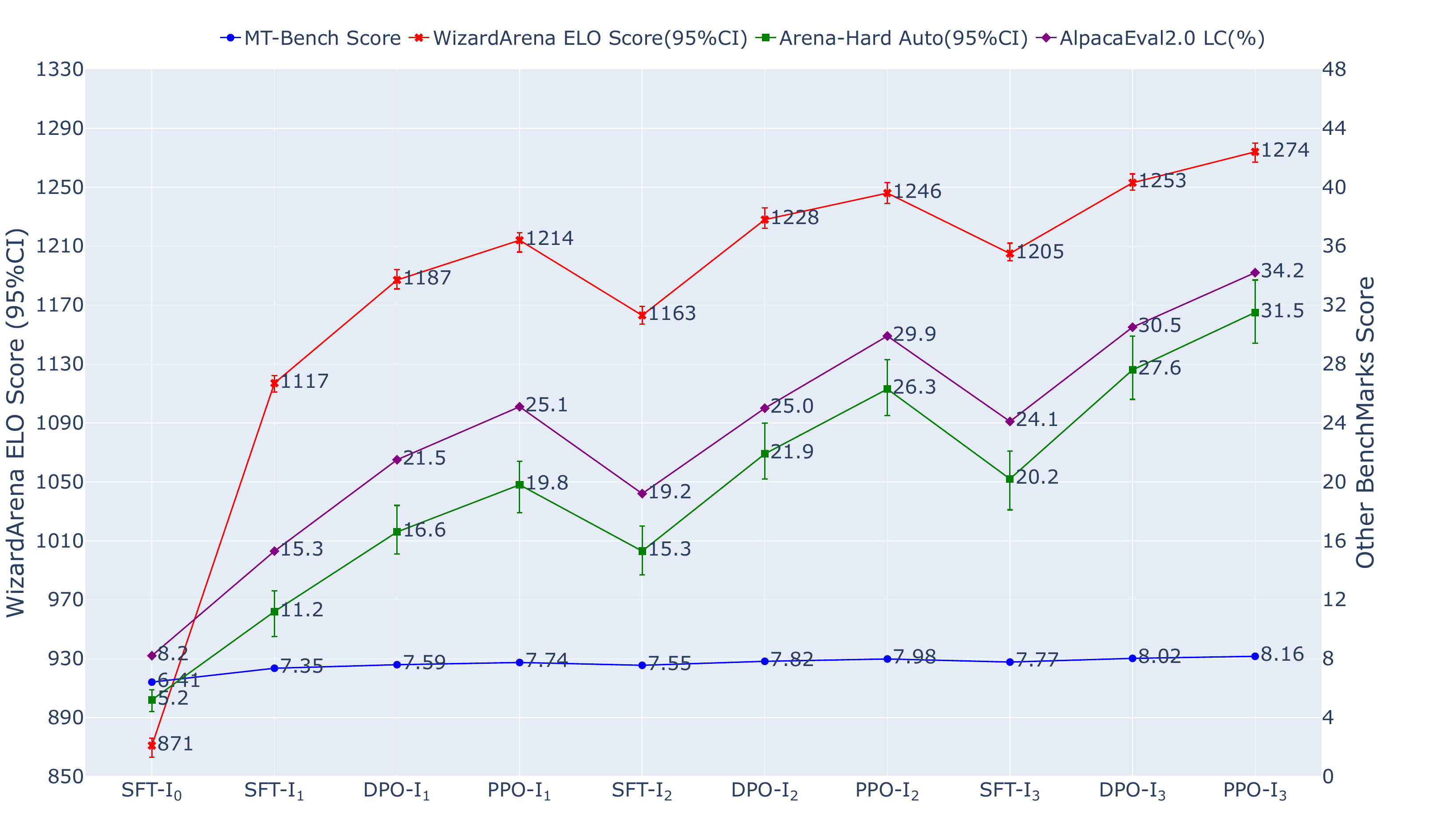}
    \caption{Explore the impact of iterative training processes of SFT, DPO, and PPO on the WizardLM-$\beta$-7B model performance in four benchmarks.}

    \label{fig:iterative_train}
    \vspace{-8pt}

\end{figure}

\subsection{Can \methodname build an effective data flywheel with post-training?}

Table ~\ref{tab: arena_elo} demonstrates the impact of using the \methodname method to post-train WizardLM-$\beta$ models during three data flywheel iterations, where \(I_i\) represents the \(i\)-th iteration. In each iteration from $I_1$ to $I_3$, we always use 90k data for post-training. Starting from WizardLM-$\beta$-7B-$I_0$, the next 3 iterations have improved by 343 points, 32 points, and 28 points on Wizardarena-Mix Elo, respectively. At the same time, the MT-bench score of this model has also achieved significant improvement (from 6.41 to 8.16). Specifically, the WizardLM-$\beta$-7B-$I_1$ even surpasses WizardLM-70B-v1.0 and the WizardLM-$\beta$-7B-$I_3$ also shows comparable performance with Starling-LM-7B-Beta. It is worth noting that we have also observed the same trend on WizardLM-$\beta$-8x22B models, and even achieved a more significant increase in both Wizardarena-Mix Elo (+460) and MT-Bench (+2.07). This model also beats both Command R+ and Claude 3 Haiku. Figure ~\ref{fig:win_rate} presents the win rates of 32 models in WizardArena-Mix, with each model involving in 2k x 31 battles. Compared to those baselines, our model has achieved significant improvements in win rate from the $I_0$ to $I_3$. Specifically, using GPT-4o as the battle target, our WizardLM-$\beta$-8x22B's win rate increased by 26\% (8\% -> 22\% -> 27\% ->34\%), WizardLM-$\beta$-7B's win rate also increased by 14\% (6\% -> 16\% -> 18\% ->20\%).

Above results highlight that continuous battle with SOTA models with \methodname and updating weights with new selected data can progressively enhance model capacities compared to its rivals. Hence, \methodname builds an effective data flywheel and utilizing the \methodname can significantly improve model performance in post-training.

\subsection{Scaling Iterative SFT, DPO, and PPO with \methodname.} 

As the core question of this paper asks how \methodname improves a model’s performance with post-training, in this section we examine how performance is affected by different post-training technology and data flywheel iterations. Figure ~\ref{fig:iterative_train} explores the results of WizardLM-$\beta$-7B model. As expected, we observe that each performance across the SFT and RL models improves step by step as we add more selected data from more \methodname battle iterations. Specifically, from SFT-\(I_0\) to PPO-\(I_3\), the WizardArena-Mix ELO score improves from 871 to 1274, achieves a huge gain of 403 points, and the Arena-Hard Auto ELO score also rises by 26.3 points (from 5.2 to 31.5). Additionally, the AlpacaEval 2.0 LC win rate improved by 26\%, from 8.2\% to 34.2\%, and the MT-Bench score increased by 1.75 points, from 6.41 to 8.16. Significant improvements across four key benchmarks highlight the effectiveness and scalability of the iterative training approach proposed by \methodname in enhancing post-training LLMs during the SFT, DPO, and PPO stages.

\newpage

\subsection{Ablation Study}

\begin{wraptable}{r}{0.5\textwidth}
\vspace{-12.5pt}
    \centering
    \caption{Explores data selection strategies during the first round of SFT stage, using 10k samples for each method except for the Original $D_1$.}
    \scalebox{0.70}{
    \setlength{\tabcolsep}{8pt}

                \begin{tabular}{lccc}
                \toprule
                Data Selection & \makecell{Data \\ Size} & \makecell{WizardArena-Mix \\ ELO (95\% CI)} & MT-Bench \\
                \midrule
                Original Data & 30k & 1079 (+5/-8) & 6.88 \\
                Random Sample & 10k & 1072 (+8/-7) & 6.77\\
                K-Means Cluster & 10k  & 1085 (+7/-5) & 6.98\\
                Instruction Length & 10k & 1081 (+5/-9) & 6.92\\
                IFD~\cite{li2023_IFD} & 10k & 1091 (+7/-6) & 7.07\\
                INSTAG~\cite{lu2023instag} & 10k & 1096 (+5/-8) & 7.12\\
                \midrule
                Pair-judge  & 10k & 1108 (+6/-8) & 7.23\\
                \bottomrule
            \end{tabular}
    \label{tab:data_selection}
}
\end{wraptable}

\textbf{Data Selection strategy.} To explore the efficiency of our pair-judge data selection method, we compare it with some widely used data selection strategies during
the first round of SFT stage. In Table ~\ref{tab:data_selection}, we use 10k samples for each method except for the Original $D_1$. The results indicate that data selected via the pair-judge method yielded a 29-point improvement in the WizardArena-Mix ELO over the all original 30k data, surpassed the diversity-based K-Means Cluster method by 23 points, and exceeded the instruction complexity-based INSTAG~\cite{lu2023instag} method by 12 points. On MT-bench, the pair-judge method also demonstrated superior performance, with improvements of 0.35 points over Original Data, 0.25 points over K-Means Cluster, and 0.11 points over INSTAG. This advantage is attributed to that the pair-judge method focuses on instructions where the base model underperforms, particularly in diverse and complex tasks, effectively addressing the model's weaknesses.  Simultaneously, these results underscore the effectiveness of the pair-judge method in selecting high-quality data during the SFT stage to target and strengthen the weakness of the base model.

\begin{figure}[h]
    \centering
    \begin{minipage}{0.485\linewidth}
        \centering
        \includegraphics[width=1\linewidth, trim=0 0 36 0,clip]{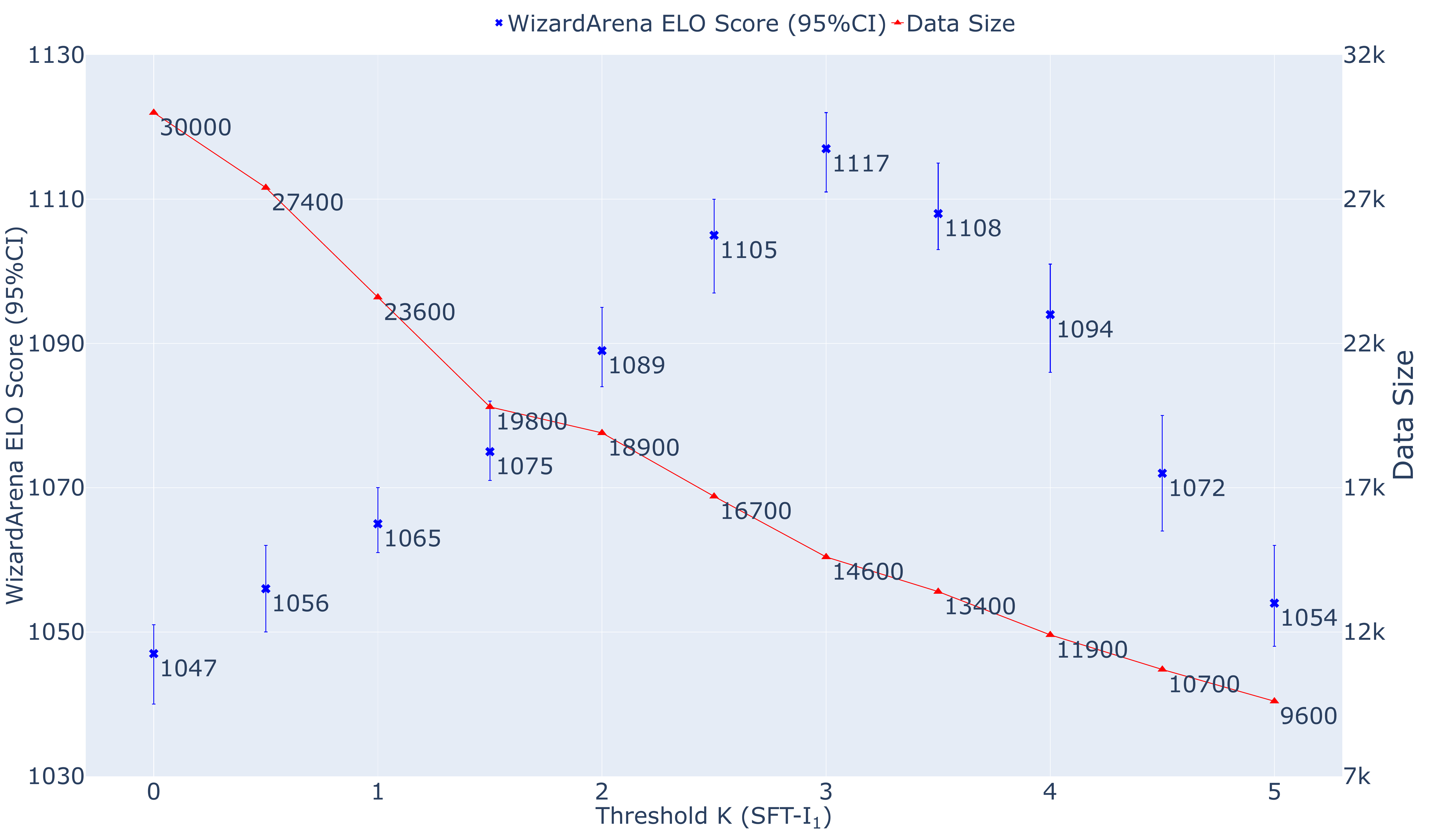}
        \label{fig:sft-k}
    \end{minipage}
    \hspace{0.01\linewidth}
    \begin{minipage}{0.485\linewidth}
        \centering
        \includegraphics[width=1\linewidth, trim=0 0 36 0,clip]{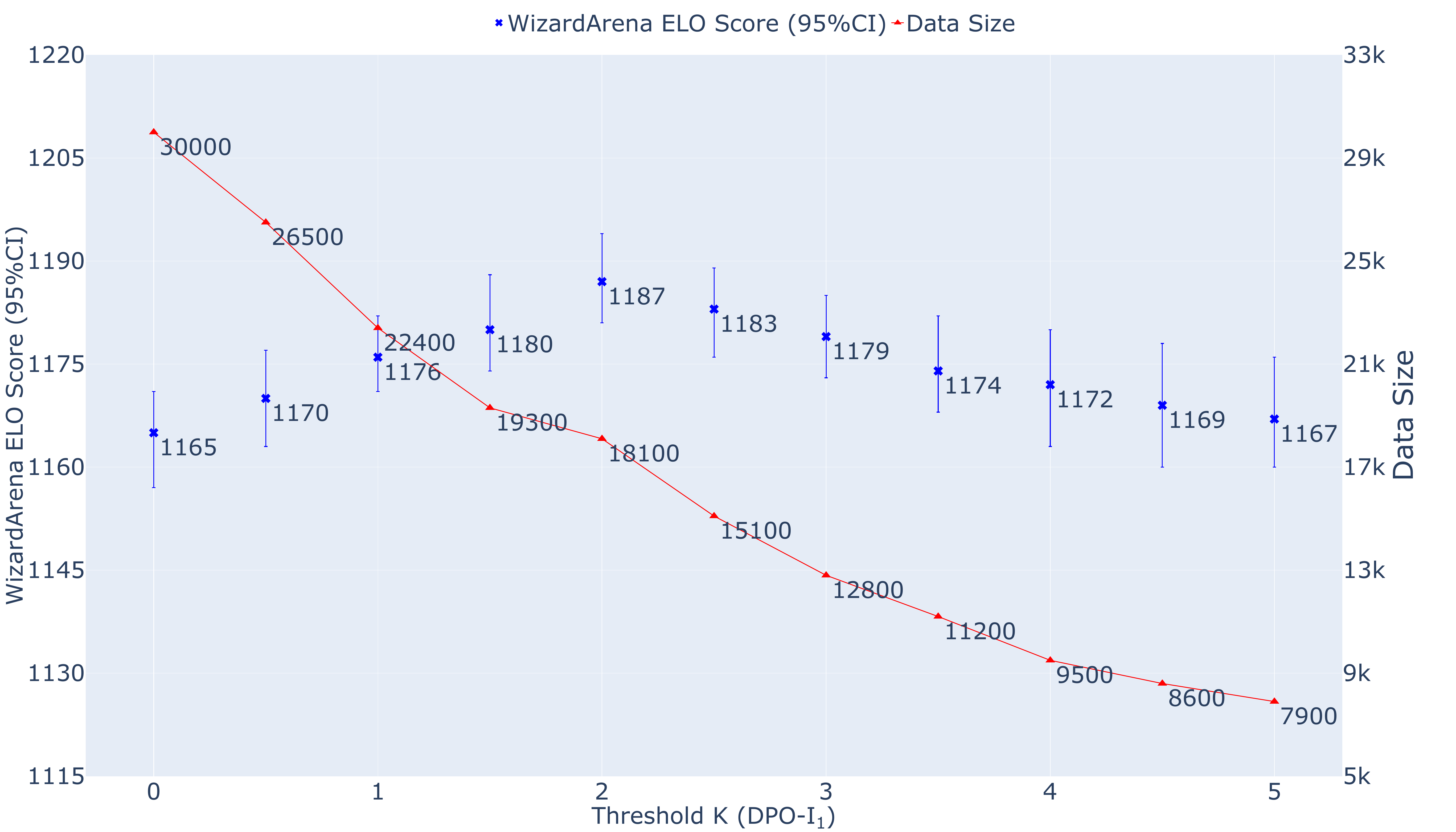}
        \label{fig:dpo-k}
    \end{minipage}
    \caption{Explore the impact of the threshold K on the WizardLM-$\beta$-7B model during the first round of SFT and DPO.}
    \label{fig:sft-dpo-k}
\end{figure}

\textbf{The relationship between data size and performance.}
An intuitive question is whether the improvement in model performance is solely due to the increase in data size. Therefore, in this section, we discuss the impact of data size and quality on model performance. Threshold is an important hyperparameter in \methodname that controls the  size of SFT data and gap between <chosen, reject> pairs of RL data. We conducted the experiments of WizardLM-$\beta$-7B-SFT-\(I_1\) and  WizardLM-$\beta$-7B-DPO-\(I_1\) where threshold ranges from 0 to 5. The result is shown in the Figure \ref{fig:sft-dpo-k}, and we did observe the best threshold of SFT and DPO data are 3.0 and 2.0 respectively in $I_1$. In SFT, compared to threshold=0, although half of the training data (30k -> 14.6k) is left when the threshold=3, the ELO of the model actually brings a 70-point improvement (1047 -> 1117). Similarly in DPO, setting the threshold=2 reduced the data to 18.1k compared to threshold=0, and the ELO of the model improved by 22 points (1165 -> 1187). This indicates that the battle helps us filter out the truly needed data, thereby constructing a more efficient data flywheel with a more streamlined scale.

\begin{table}[ht]
\vspace{-12pt}
    \centering
    \caption{Explore the consistency between Llama3-70B-Instruct and GPT-4 as judging models in the Offline-Mix Arena. Using multiple bootstraps (i.e., 100), we select the median as the model's ELO score and employ Llama-2-70B-Chat ELO score as the reference point.}
    \scalebox{0.65}{
    \begin{tabular}{llcccc}
    \toprule
    Model & \makecell{LMSYS-ChatBot \\ Arena-ELO-EN \\ (95\% CI)} & \makecell{WizardArena-Mix-ELO \\ GPT-4-judge \\ (95\% CI)} & \makecell{WizardArena-Mix-ELO \\ Llama3-70B-Instruct-judge \\ (95\% CI)} & \makecell{WizardArena-Mix-ELO \\ \{GPT-4 \& Llama3-70B-Instruct\}-judge \\ (95\% CI)} \\
    \midrule
    GPT-4o~\cite{openai2023gpt4} & 1266 (+4/-4) & 1388 (+5/-3) & 1395 (+5/-4) & 1399 (+5/-4)  \\
    Calude 3.5 Sonnet~\cite{anthropic2024claude} & 1246 (+4/-7) & 1372 (+6/-6) & 1384 (+6/-4) & 1387 (+6/-6)\\
    Gemini 1.5 Pro~\cite{team2023gemini} & 1235 (+5/-4) & 1365 (+4/-3) & 1377 (+5/-5) & 1375 (+5/-5)\\
    Command R+~\cite{cohere} & 1163 (+4/-4) & 1349 (+5/-7) & 1337 (+6/-4)  & 1340 (+4/-4)\\
    Claude 3 Haiku~\cite{anthropic2024claude} & 1158 (+4/-3) & 1355 (+3/-5) & 1342 (+4/-6)  & 1346 (+3/-4)\\
    Qwen1.5-72B-Chat~\cite{Bai2023QwenTR} & 1135 (+3/-4) & 1331 (+6/-5) & 1321 (+6/-5) & 1327 (+5/-5)\\
    Qwen1.5-32B-Chat~\cite{Bai2023QwenTR} & 1109 (+4/-5) & 1297 (+4/-7) & 1283 (+6/-4)  & 1278 (+7/-4)\\
    Starling-LM-7B-Beta~\cite{zhu2023starling} & 1108 (+5/-5) & 1275 (+6/-7) & 1272 (+4/-6)  & 1274 (+5/-5)\\
    WizardLM-70B-v1.0~\cite{xu2023wizardlm} & 1098 (+7/-6) & 1107 (+5/-4) & 1169 (+5/-5)  & 1166 (+6/-4)\\
    LLama-2-70B-Chat~\cite{touvron2023llama2} & 1097 (+5/-4) & 1100 (+0/-0) & 1100 (+0/-0)  & 1100 (+0/-0) \\
    Nous-Hermes-2-Mixtral-DPO~\cite{Nous-Hermes-2-Mixtral-8x7B-DPO} & 1078 (+9/-8) & 1063 (+7/-8) & 1114 (+5/-4)  & 1109 (+7/-8)\\
    DeepSeek-LLM-67B-Chat~\cite{Bi2024DeepSeekLS} & 1065 (+12/-10) & 985 (+7/-9) & 1000 (+7/-5)  & 998 (+4/-7)\\
    Llama-2-13B-Chat~\cite{touvron2023llama2} & 1061 (+5/-6) & 974 (+7/-5) & 1042 (+5/-4)  & 1044 (+6/-6)\\
    GPT-3.5-Turbo-0613~\cite{openai2023gpt4} & 1052 (+5/-5) & 942 (+8/-6) & 981 (+6/-5) & 977 (+7/-6)\\
    Zephyr-7b-alpha~\cite{Tunstall2023ZephyrDD} & 1040 (+17/-13) & 925 (+5/-6) & 939 (+4/-5)  & 937 (+4/-5)\\
    Vicuna-13B~\cite{vicuna2023} & 1029 (+6/-5) & 939 (+5/-8) & 927 (+5/-5) & 927 (+6/-6)\\
    Qwen-14B-Chat~\cite{Bai2023QwenTR} & 1017 (+9/-10) & 916 (+6/-6) & 924 (+4/-6) & 923 (+4/-6)\\
    \bottomrule
    \end{tabular}
    }
    \vspace{-8pt}
    \label{tab:judge_models}
    
\end{table}

\textbf{Llama3-Chat Judge or GPT-4 Judge?}
In most previous works, people were accustomed to use GPT-4 as a judge for evaluation or generating synthetic data, but the GPT-4 API cost required for large-scale data flywheel is enormous for most research and production scenarios. Therefore, we explore whether it is possible to replace GPT-4 with advanced open source models. Table ~\ref{tab:judge_models} explores the consistency between Llama3-70B-Instruct and GPT-4 as judge models in the WizardArena-Mix Arena. Using GPT-4 judge's ELO as the reference benchmark, the Spearman correlation coefficient between Llama3-70B-Instruct judge and GPT-4 judge  is 99.26\%, and the Human Agreement with 95\% CI is 96.15\%. The overall average consistency between the two judge models is 97.71\%. Furthermore, combining GPT-4 and Llama3-70B-Instruct as the judge model resulted in an overall average consistency of 98.40\% for LMSYS ChatBot Arena, a slight 0.25\% improvement over using only Llama3-70B-Instruct (98.40\% vs. 98.15\%). Consequently, employing Llama3-70B-Instruct as a cost-effective judge model achieves high consistency with both GPT-4 and LMSYS ChatBot Arena by human judgment, ensuring the reliability of the WizardArena evaluation and post-training with \emph{Arena Learning} in this paper.

\begin{wrapfigure}{r}{0.50\textwidth}
\vspace{-12pt}
    \centering
    \includegraphics[width=1\linewidth]{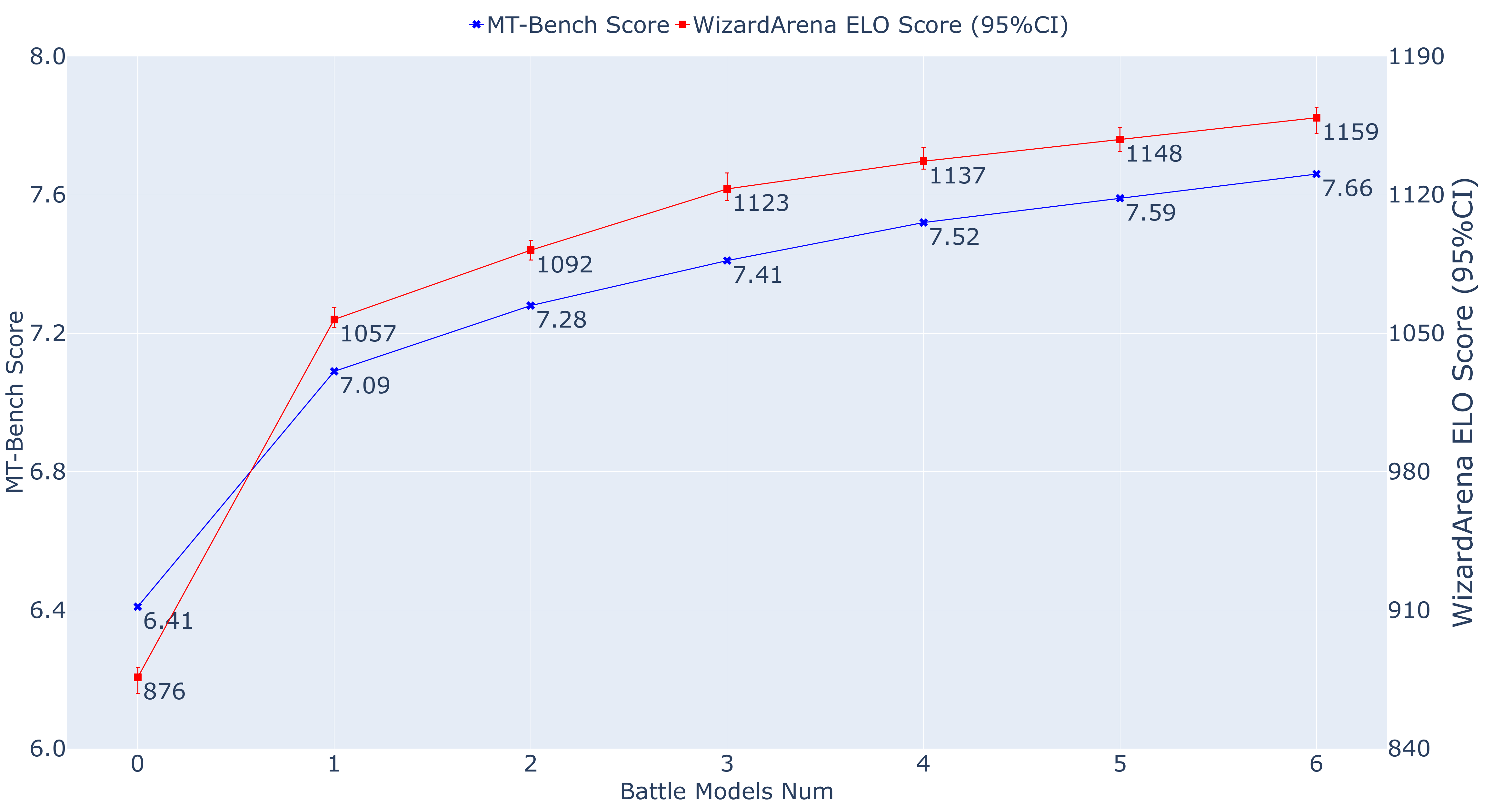}
    \caption{Explore the impact of the scale of battle models on  WizardLM-$\beta$-7B-SFT-\(I_1\).}
    \label{fig:scale_battle_models}
    \vspace{-12pt}
\end{wrapfigure}

\textbf{Number of battle models.} Figure \ref{fig:scale_battle_models} presents an ablation study investigating the impact of the number of other battle models. According to Table ~\ref{tab: arena_elo}, the models are ranked in descending order based on WizardArena-Mix ELO scores. Subsequently, models ranging from Command R+ to OpenChat 3.5 are selected for battle. As the number of models participating in the battle increases, the performance of the WizardLM-$\beta$-7B-SFT-\(I_1\) model gradually increases. Specifically, on WizardArena-Mix, the ELO rating of WizardLM-$\beta$-7B increases from 876 to 1159, a gain of 283 points. Concurrently, the MT-Bench score rises from 6.41 to 7.66, an increase of 1.25 points. This demonstrates the scalability of our method and its compatibility with different models, providing a basis for future large-scale application of \methodname. However, as relationship between the complexity of the battle $O(\cdot)$ and the number of models $n$ is $O(n^2)$, and in order to balance the computational cost and model performance, we chose 3 other models to battle with WizardLM-$\beta$ as the default setting in this paper.

\begin{wraptable}{r}{0.5\textwidth}
\vspace{-12.5pt}
    \centering
    \caption{The WizardArena Elo of WizardLM-$\beta$-7B-SFT-\(I_1\) on different battle modes.}
    \scalebox{0.62}{
    \setlength{\tabcolsep}{1pt}
\begin{tabular}{l|c}
            \toprule
            Battle Mode & WizardArena  \\
            \midrule
            i) Ours v.s. OpenChat-3.5 & 924 (+7/-5)  \\       
            i) Ours v.s. Qwen-1.5-72B & 1015 (+5/-5)  \\      
            i) Ours v.s. Command R+ & 1028 (+6/-4) \\
            ii) Ours v.s. \{Qwen-1.5-72B/OpenChat-3.5/Command R+\} & 1046 (+5/-8) \\  
            iii) \{Ours, Qwen-1.5-72B, OpenChat-3.5\}, 1v.s.1 & 1052 (+6/-7)   \\ 
            iii) \{Ours, Command R+, OpenChat-3.5\}, 1v.s.1 &  1065 (+5/-8)\\ 
            iii) \{Ours, Qwen-1.5-72B, Command R+\}, 1v.s.1 &  1095 (+5/-5) \\ 
            iv) \{Ours, Qwen-1.5-72B, Command R+, OpenChat-3.5\}, 1v.s.1 & 1117 (+5/-6) \\ 
            \bottomrule
        \end{tabular}
    \label{tab:diff_battle_modes_1}
}
\end{wraptable}
\textbf{The impact of different battle modes.} In order to explore the necessity of using multiple models pairwise battle  to construct a data flywheel, we designed various battle modes on $D_1$ SFT data, including: i) \{ours + 1 other model\} pairwise battle with each other, ii) randomly split $D_1$ into 3 parts, ours battle with one other model on each part respectively, iii) \{ours + 2 other models\} pairwise battle with each other, iv)  \{ours + 3 other models\} pairwise battle with each other. We use WizardLM-$\beta$-7B-SFT-\(I_0\), Openchat-3.5, Qwen-1.5-72B, and CommandR+ as the battle group in this section, the output model is WizardLM-$\beta$-7B-SFT-\(I_1\). As shown in the Table \ref{tab:diff_battle_modes_1}, the mode (iv) achieved best performance on WizardArena and Outperformed the (i) mode \{Only Command R+ battle\} by 89 points and the (iii) mode \{Command R+ \& Qwen1.5-72B-Chat Battle\}  by 22 points. To this end, we finally leverage multiple models pairwise battle with each other to build the simulated offline Chatbot Arena.

\begin{table}[h]
\vspace{-12pt}

\centering
\caption{Explore the performance of the WizardLM-$\beta$ model across various benchmarks. The results of baselines are cited from Arena-Hard Auto~\cite{arenahard2024_lmsysarena_hard}, AlpacaEval 2.0 LC ~\cite{alpaca_eval}, and OpenLLM Leaderboard~\cite{open-llm-leaderboard}. }

\scalebox{0.68}{
\begin{tabular}{l|l|c|ccccc}
\toprule
\textbf{Model} & \makecell{\textbf{Arena-Hard Auto} \\ \textbf{(95\% CI)}} & \makecell{\textbf{AlpacaEval 2.0 LC} \\ \textbf{(Win Rate \%)} } & \textbf{ARC} & \textbf{Hellaswag} & \textbf{MMLU} & \textbf{TruthfulQA} & \textbf{Avg.} \\ 
\midrule
Claude 3.5 Sonnet~\cite{anthropic2024claude} & 79.3 (-2.1, 2.0) & 52.4 & - & - & - & - & - \\
GPT-4o~\cite{openai2023gpt4} & 79.2 (-1.9, 1.7) & 57.5 & - & - & - & - & - \\
GPT-4-0125-Preview~\cite{openai2023gpt4} & 78.0 (-2.1, 2.4) & - & - & - & - & - & - \\
Gemini 1.5 Pro~\cite{team2023gemini} & 72.0 (-2.1, 2.5) & - & - & - & - & - & - \\
WizardLM-2-8x22B-0415~\cite{xu2023wizardlm} & 69.6 (-1.8, 2.4) & 51.3 & - & - & - & - & - \\
GLM-4-0520~\cite{GLM} & 63.8 (-2.9, 2.8) & - & - & - & - & - & - \\
Yi-Large~\cite{young2024yi} & 63.7 (-2.6, 2.4) & 51.9 & - & - & - & - & - \\
DeepSeek-Coder-V2-Instruct~\cite{zhu2024deepseek_coder} & 62.3 (-2.1, 1.8) & - & - & - & - & - & - \\
Gemma-2-27B-it~\cite{team2024gemma} & 57.5 (-2.1, 2.4) & - & - & - & - & - & - \\
GPT-4-0314~\cite{openai2023gpt4} & 50.0 (0.0, 0.0) & 35.3 & - & - & - & - & - \\
Qwen2-72B-Instruct~\cite{Bai2023QwenTR} & 46.9 (-2.5, 2.7) & - & - & - & - & - & - \\
Claude 3 Sonnet\cite{anthropic2024claude} & 46.8 (-2.3, 2.7) & 34.9 & - & - & - & - & - \\
Llama-3-70B-Instruct~\cite{touvron2023llama2} & 41.1 (-2.0, 2.2) & 34.4 & 71.42 & 85.69 & 80.06 & 61.81 & 74.75 \\
Mixtral-8x22b-Instruct-v0.1~\cite{jiang2023mistral} & 36.4 (-2.4, 2.6) & 30.9 & 72.70 & 89.08 & 77.77 & 68.14 & 76.92 \\
Qwen1.5-72B-Chat~\cite{Bai2023QwenTR} & 36.1 (-2.0, 2.7) & 36.6 & 68.26 & 86.47 & 77.46 & 63.84 & 74.01 \\
Phi-3-Medium-4k-Instruct~\cite{abdin2024phi} & 33.4 (-2.6, 2.1) & - & 67.32 & 85.76 & 77.83 & 57.71 & 72.16 \\
Command R+~\cite{cohere} & 33.1 (-2.8, 2.4) & - & 70.99 & 88.56 & 75.73 & 56.30 & 72.90 \\
GPT-3.5-Turbo-0613~\cite{openai2023gpt4} & 24.8 (-1.9, 2.3) & 22.7 & - & - & - & - & - \\
DBRX-Instruct~\cite{DBRX} & 23.9 (-1.5, 1.5) & 25.4 & 67.83 & 88.85 & 73.72 & 67.02 & 74.36 \\
Yi-34B-Chat~\cite{young2024yi} & 23.1 (-1.6, 1.8) & 27.2 & 70.48 & 85.97 & 77.08 & 62.16 & 73.92 \\
Phi-3.1-Mini-4k-Instruct~\cite{abdin2024phi} & 23.1 (-2.4, 2.0) & - & 62.97 & 80.6 & 69.08 & 59.88 & 68.13 \\
Starling-LM-7B-Beta~\cite{zhu2023starling} & 23.0 (-1.8, 1.8) & - & 67.24 & 83.47 & 65.14 & 55.47 & 67.83 \\
Llama-3-8B-Instruct~\cite{touvron2023llama2} & 20.6 (-2.0, 1.9) & 22.9 & 60.75 & 78.55 & 67.07 & 51.65 & 64.51 \\
Tulu-2-DPO-70B~\cite{ivison2023_tulu2} & 15.0 (-1.6, 1.3) & 21.2 & 72.10 & 88.99 & 69.84 & 65.78 & 74.18 \\
Mistral-7B-Instruct-v0.1~\cite{jiang2023mistral} & 12.6 (-1.7, 1.4) & - & 54.52 & 75.63 & 55.38 & 56.28 & 60.45 \\
Llama-2-70B-Chat~\cite{touvron2023llama2} & 11.6 (-1.5, 1.2) & 14.7 & 64.59 & 85.88 & 63.91 & 52.80 & 66.80 \\
Vicuna-33B~\cite{vicuna2023} & 8.6 (-1.1, 1.1) & 17.6 & 62.12 & 83.00 & 59.22 & 56.16 & 65.13 \\
Gemma-7B-it~\cite{team2024gemma} & 7.6 (-1.2, 1.3) & 10.4 & 51.45 & 71.96 & 53.52 & 47.29 & 56.06 \\
Llama-2-7b-chat~\cite{touvron2023llama2} & 4.6 (-0.8, 0.8) & 5.4 & 52.90 & 78.55 & 48.32 & 45.57 & 56.34 \\
Nous-Hermes-2-Mixtral-DPO~\cite{Nous-Hermes-2-Mixtral-8x7B-DPO} & - & - & 71.42 & 87.21 & 72.28 & 54.53 & 71.36 \\
DeepSeek-LLM-67B-Chat~\cite{Bi2024DeepSeekLS} & - & 17.8 & 67.75 & 86.8 & 72.19 & 55.83 & 70.64 \\
OpenChat-3.5-0106~\cite{wang2023openchat} & - & - & 66.04 & 82.93 & 65.04 & 51.90 & 66.48 \\
Zephyr-7b-beta~\cite{Tunstall2023ZephyrDD} & - & 13.2 & 62.03 & 84.36 & 61.07 & 57.45 & 66.23 \\
Qwen1.5-7B-Chat~\cite{Bai2023QwenTR} & - & 14.7 & 55.89 & 78.56 & 61.65 & 53.54 & 62.41 \\
Vicuna-13b-v1.5~\cite{vicuna2023} & - & 11.7 & 57.08 & 81.24 & 56.67 & 51.51 & 61.63 \\
Llama-2-13B-Chat~\cite{touvron2023llama2} & - & 8.4 & 59.04 & 81.94 & 54.64 & 44.12 & 59.94 \\
\midrule
WizardLM-$\beta$-7B--$I_0$ & 5.2 (-0.8, 0.7) & 8.2 & 54.73 & 72.67 & 54.43 & 49.16 & 57.75 \\
WizardLM-$\beta$-7B--$I_1$ & 19.8 (-1.9, 1.6) & 25.1 & 60.32 & 83.11 & 61.50 & 55.92 & 65.21 \\
WizardLM-$\beta$-7B--$I_2$ & 26.3 (-1.8, 2.0) & 29.9 & 62.25 & 84.38 & 63.96 & 56.67 & 66.82 \\
WizardLM-$\beta$-7B--$I_3$ & 31.5 (-2.1, 2.2) & 34.2 & 64.58 & 84.93 & 65.74 & 57.06 & 68.08 \\
WizardLM-$\beta$-8x22B-$I_3$ & 64.3 (-2.0, 2.5) & 48.9 & 67.91 & 86.64 & 73.76 & 66.48 & 73.70 \\
\bottomrule
\end{tabular}
}
\vspace{-12pt}

\label{tab:other_benchmarks}
\end{table}

\textbf{Performance on more benchmarks.}
Table ~\ref{tab:other_benchmarks} highlights the performance of WizardLM-$\beta$ across various metrics after three iterations, including LMSYS Arena-Hard Auto, AlpacaEval 2.0 LC, and the OpenLLM Leaderboard. In LMSYS Arena-Hard Auto, WizardLM-$\beta$-7B's score rises from 5.2 to 31.5, with a gain of 26.3 points, surpassing GPT-3.5-Turbo-0613 by 6.7 points and Llama 3-8B-Instruct by 10.9 points, closely aligning with Command R+. WizardLM-$\beta$-8x22B's performance outperforms Llama-3-70B-Instruct by 23.2 points, is also better than GLM-4-0520 and Yi-Large. In AlpacaEval 2.0 LC, WizardLM-$\beta$-7B's win rate increases from 8.2\% to 34.2\%, exceeding GPT-3.5-Turbo-0613 by 11.5 points and Mixtral-8x22b-Instruct-v0.1 by 3.3 points, matching closely with Llama3-70B-Instruct. Moreover, WizardLM-$\beta$-8x22B's win rate even surpasses Llama-3-70B-Instruct by 14.5 points and GPT-4-0314 by 13.6 points. On the OpenLLM Leaderboard, WizardLM-$\beta$-7B's average score increases from 57.75 to 68.08, surpassing Llama-2-70B-Chat by 1.28 points and comparable to Starling-LM-7B-beta. WizardLM-$\beta$-8x22B is also compareable with Command R+, exceeds Deepseek-LLM-67B-Chat by 3.06 points, and closely approaches Qwen1.5-72B-Chat and Llama-3-70B-Instruct. The above results indicate that: 1) Utilizing the \methodname method to generate training data significantly improves the performance of the model by multiple training iterations. 2) \methodname  can improves the generalization and scalability of the model performance.

\begin{wraptable}{r}{0.5\textwidth}
\vspace{-12.5pt}
    \centering
    \caption{Data count and difficulty of each iteration.}
    \scalebox{0.8}{
\begin{tabular}{l|c|c|c}
            \toprule
             & Threshold & Count  & Difficulty  \\ \midrule
             Original & - & 30k x 3 & 4.7  \\ \midrule
            SFT-\(I_1\) & 3.0 & 14.6k & 5.8  \\
            SFT-\(I_2\) & 1.0 & 11.3k & 6.5  \\
            SFT-\(I_3\) & 1.0 & 7.8k & 7.4  \\ \midrule
            SFT-Total & - & 33.7k & 6.4  \\
            \bottomrule
        \end{tabular}
    \vspace{-1cm}
    \label{tab:data_count_and_difficulty_2}
}
\end{wraptable}

\textbf{Data count and difficulty of each iteration.}
In table ~\ref{tab:data_count_and_difficulty_2} we show in detail the data size, difficulty, and threshold division for each round of the SFT.
As the number of iteration rounds increased, we adjusted the threshold from 3 to 1, but the data size of SFT still significantly decreased (30k -> 7.8k). This is because as the model's ability evolved, the number of battles it lost also sharply declined. We also found that the difficulty of each round of data gradually increases (4.7 -> 7.4) and we only need totally around 1/3 data for final SFT (90k -> 33.7k) and the average difficulty is 6.4. It indicates that a reasonable data flywheel should focus more on finding those challenging data for target model to fill in the shortcomings of its capabilities.

\begin{table}[h]
\vspace{-8pt}

    \centering
    \caption{Explore the quantity of selected responses for each battle model across various rounds during the SFT and DPO stages.}
    \scalebox{0.75}{
    \setlength{\tabcolsep}{10pt}
    \begin{tabular}{l|c|c|c|c|c}
        \toprule
        Stage & Command R+ & Qwen1.5-72B-Chat & OpenChat-3.5 & WizardLM-$\beta$-7B & Total \\
        \midrule
        SFT-\(I_1\) & 6.9k & 5.5k & 2.2k & - & 14.6k \\
        SFT-\(I_2\) & 5.8k & 4.2k & 1.3k & - & 11.3k \\
        SFT-\(I_3\) & 4.1k & 3.0k & 0.7k & - & 7.8k \\
        SFT-\(Total\) & 16.8k & 12.7k & 4.2k & - & 33.7k \\
        \midrule
        DPO-\(I_1\) & 8.7k & 7.6k & 1.9k & 1.1k & 19.3k \\
        DPO-\(I_2\) & 8.0k & 7.2k & 1.1k & 1.6k & 17.9k \\
        DPO-\(I_3\) & 7.4k & 6.5k & 0.6k & 2.3k & 16.8k \\
        DPO-\(Total\) & 24.1k & 21.3k & 3.6k & 5.0k & 54.0k \\
        \bottomrule
    \end{tabular}
    }
    \label{tab:stage_data}
\vspace{-8pt}
\end{table}

\textbf{Count of data selected from each battle model.}
Table ~\ref{tab:stage_data} illustrates the count of selected win/accepted responses from each battle model across 3 rounds within the SFT and DPO stages. During the SFT stages, data volume consistently declines through successive iteration rounds (14.6k -> 7.8k). Moreover, the volume of selected data strong correlates with battle model performance. For instance,  Command R+ consistently requires more data than both Qwen1.5-72B-Chat and OpenChat-3.5 (16.8k > 12.7k > 4.2k). During DPO, most other battle models always show a decreasing trend in selected data per iteration round, except for WizardLM-$\beta$, which experienced an increase in data volume (1.1k -> 1.6k -> 2.3k), this is mainly because as our model performance  improves, the proportion of its recovery in positive samples also increases gradually.

\begin{wrapfigure}{r}{0.5\textwidth}
\vspace{-12pt}
    \centering
    \includegraphics[width=1\linewidth]{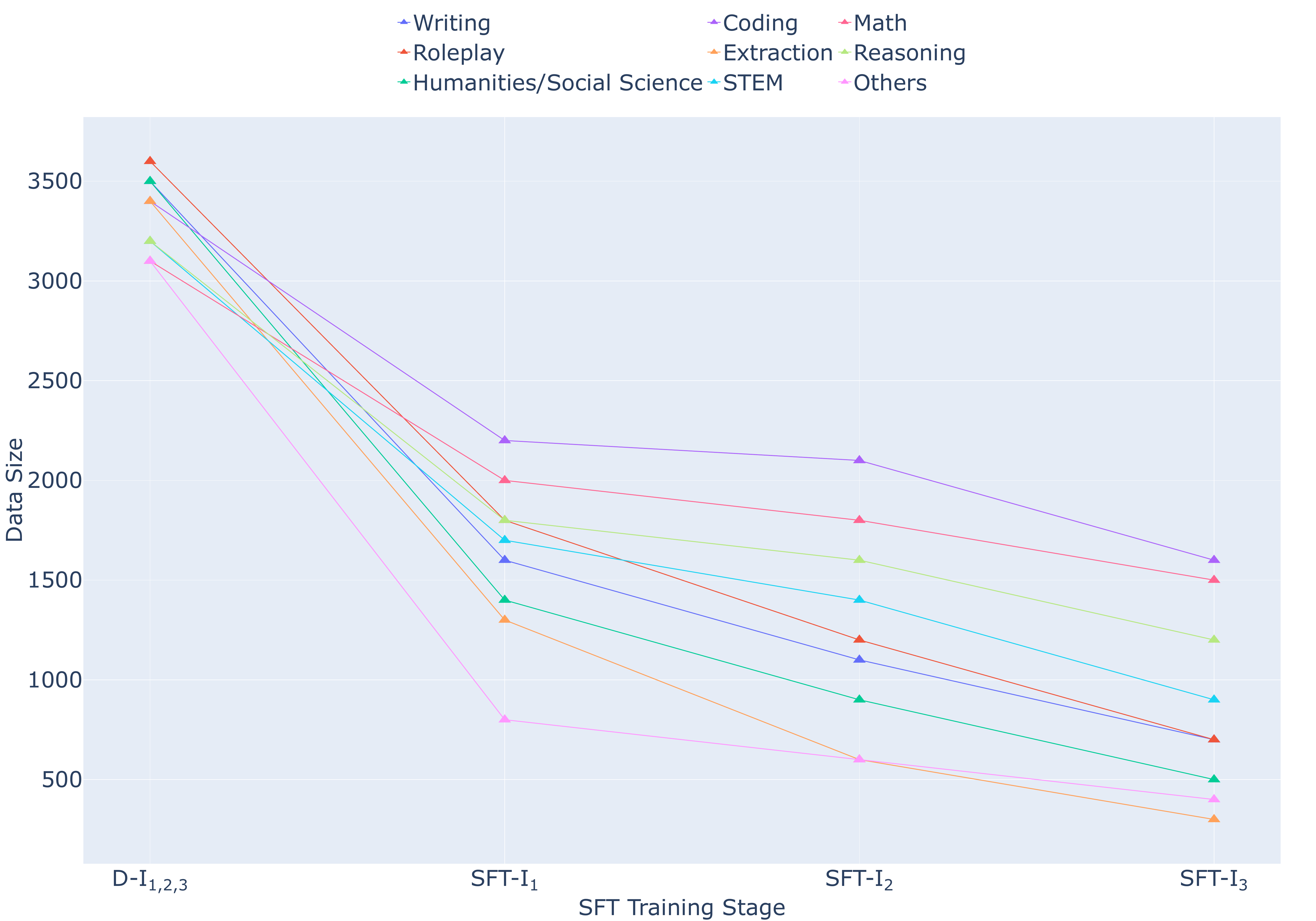}
    \caption{The selected training data size trend for SFT across each category during each iteration.}
    \label{fig:sft_data_statistics}
\end{wrapfigure}
\textbf{Data category count of each iteration.}
Figure ~\ref{fig:sft_data_statistics} illustrates the selected training data size trend for SFT across various categories during each iteration. As iterations progress, there is a consistent decline in selection across all categories. However, this decline occurs more gradually in complex categories (i.e., Mathematics, Reasoning, and Coding) while it is more pronounced in simpler categories like Writing and Extraction. Specifically, by the third iteration, the proportion of selections from more challenging categories like Coding, Math, and Reasoning has increased, whereas it has decreased for less demanding categories such as Writing and Roleplay. This pattern suggests that the selection of data progressively favors more complex tasks with each iteration, thereby significantly improving the model's performance in these intricate categories.

\begin{figure}
\vspace{-8pt}

    \centering
    \includegraphics[width=1\linewidth]{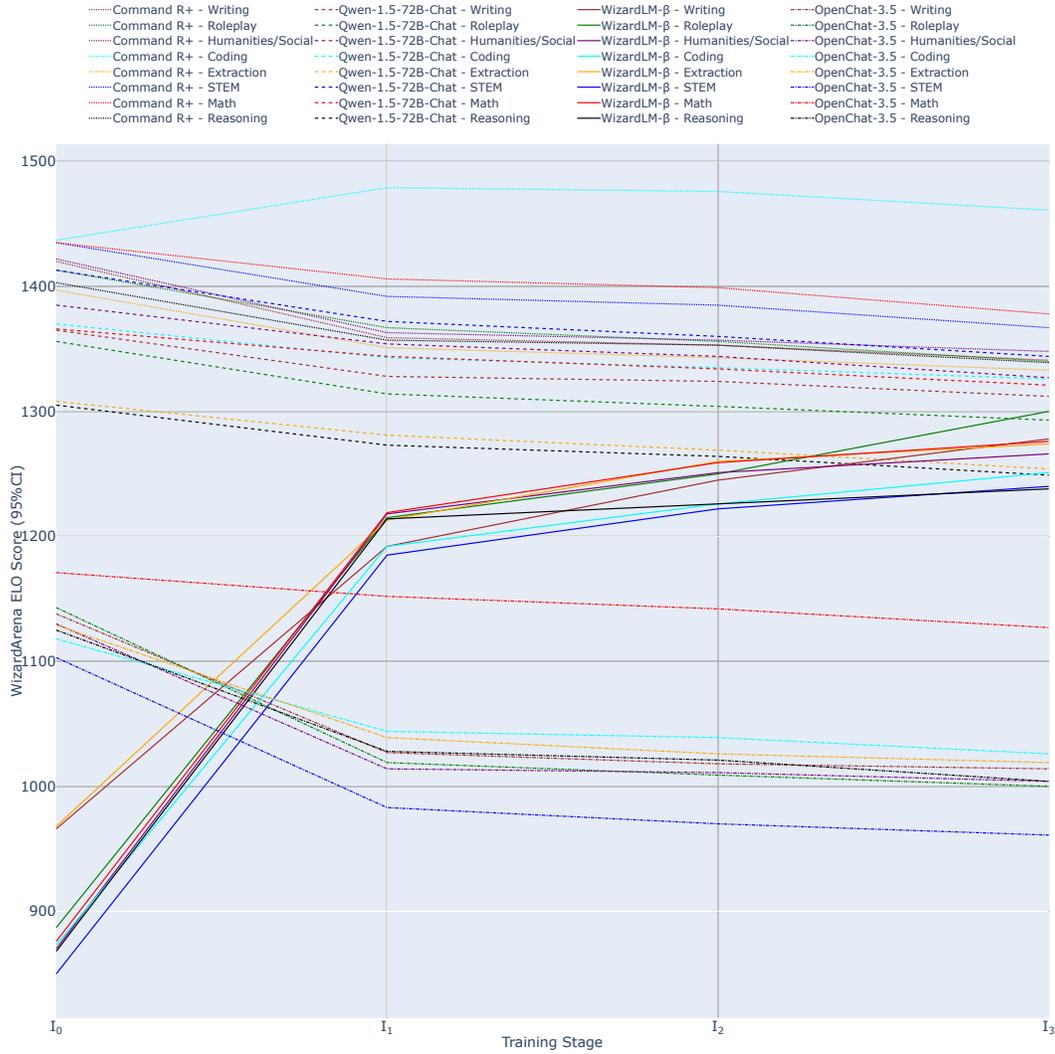}
    \caption{Explore the progression of ELO scores for the WizardLM-$\beta$-7B model across eight categories as iterations increase.}
\vspace{-8pt}
    \label{fig:category_stage_elo}
\end{figure}

\textbf{Model performance changes of each category.}
Figure ~\ref{fig:category_stage_elo} illustrates the evolution of ELO scores for the WizardLM-$\beta$-7B model across eight categories with increasing iterations during the training stage. Initially, the ELO score of WizardLM-$\beta$-7B is inferior to OpenChat 3.5. After multiple iterations, WizardLM-$\beta$-7B not only surpasses OpenChat-3.5 but also consistently approaches the performance of Qwen1.5-72B-Chat and Command R+. From iterations \(I_0\) to \(I_3\), the ELO scores of the model improve sharply across all categories, followed by a steady growth, indicating its gradual evolution from a weaker model to a stronger model. Particularly,  in less challenging categories (i.e., Roleplay and Extraction), WizardLM-$\beta$-7B begins behind but eventually outperforms Qwen1.5-72B-Chat. Conversely, in more complex reasoning tasks like Math and Coding, its progress is slower. Moreover, the  ELO battle results highlight the distinct strengths of each model. For instance, Command R+ excels in the challenging categories like Coding and Math. Meanwhile, Qwen1.5-72B-Chat shows stronger performance in Humanities/Social Science and STEM, while OpenChat3.5 is comparatively weaker. As iterations increase, training data shifts towards more complex data (i.e., Coding and Math), enhancing the model initial weaknesses. Over three rounds of iterations, our model can scale up with an extensive amount of battle training data from WizardArena, leading to substantial performance improvements. These findings highlight the significant advantages and potential of \methodname to boost post-training performance of WizardLM-$\beta$-7B by harnessing the collective knowledge and capabilities of multiple advanced models.

\newpage

\begin{table}[ht!]
\vspace{-8pt}
    \centering
    \caption{Explore the performance impact of employing more advanced models to battle with WizardLM-$\beta$-7B-\(I_0\) on different stages.}
    \scalebox{0.8}{
    \setlength{\tabcolsep}{15pt}

    \begin{tabular}{lcc}
        \toprule
         Training Stage & WizardArena Elo & MT-Bench \\ 
        \midrule
        SFT-\(I_0\) & 871 (+5/-8) &  6.41 \\
        \midrule
        \multicolumn{3}{c}{Battles With $M_0$=\{Command R+, Qwen1.5-72B-Chat, and OpenChat 3.5\}} \\
        \midrule
        SFT-\(I_1\) & 1117 (+5/-6) &  7.35 \\       
        \{SFT + DPO\}-\(I_1\) & 1187 (+7/-6)  & 7.59\\      
        \{SFT + DPO + PPO\}-\(I_1\) & 1214 (+5/-8)  & 7.74\\
        \midrule
        \multicolumn{3}{c}{Battles With $M_1$=\{GPT-4o, GPT4-1106-Preview, and WizardLM-2-8x22B\}} \\
        \midrule
        SFT-\(I_1\) & 1164 (+4/-7) &  7.60 \\       
        \{SFT + DPO\}-\(I_1\) & 1232 (+6/-6)  & 7.78\\      
        \{SFT + DPO + PPO\}-\(I_1\) & 1266 (+6/-4)  & 7.89\\
        \bottomrule
    \end{tabular}
    \label{tab:battle_top_model}
}
\end{table}

\textbf{Learning from more advanced models.} Table ~\ref{tab:battle_top_model} analyzes the performance impact of employing more advanced models to battle for WizardLM-$\beta$-7B. Initially, leveraging the $M_1$ models =\{GPT-4o, GPT-4 Turbo, and WizardLM-2-8x22B\}  in the first round improve the ELO score from the baseline SFT-\(I_0\) of 871 to 1266, a gain of 395 points and represent a 52-point improvement over batting with the $M_0$ models=\{Command R+, Qwen1.5-72B-Chat, and OpenChat 3.5\} . Throughout various stages of the battle and training, the ELO scores using the $M_1$ models are always correspondingly 45 \textasciitilde 55 points higher than the $M_0$ models. Additionally, the MT-Bench score increased from 6.41 to 7.89, marking a 0.15 point advance over $M_0$ models score of 7.74. The results highlight the substantial performance improvements that can be achieved by employing more advanced models for battle.

\section{Related Works}

\subsection{Large Language Models}
LLMs have made significant strides in Natural Language Processing (NLP), serving as a versatile foundation for numerous applications~\cite{zhao2023_llmsurvey,wang2023_llm_aligning, kopf2024openassistant_llm_align2}. These models, which often contain hundreds of billions of parameters, are trained on expansive text datasets. Notable examples include OpenAI's GPT-3 and GPT-4~\cite{GPT3,openai2023gpt4}, Anthropic's Claude~\cite{bai2022constitutional-claude}, Google's PaLM~\cite{PaLM,palm2}, Gemini~\cite{team2023gemini}, Gemma~\cite{team2024gemma}, and DeepMind's Chinchilla~\cite{Chinchilla}.
The AI field has recently seen a surge in open-source LLMs, providing public access to model codes and parameters. Notable releases include BigScience's BLOOM~\cite{workshop2023bloom}, Mistral AI's Mistral~\cite{jiang2023mistral}, Microsoft's Phi~\cite{abdin2024phi}, Meta's Llama family~\cite{touvron2023llama,touvron2023llama2, Llama3} and GAL~\cite{taylor2022galactica},  NVIDIA's Nemotron-4 340B~\cite{adler2024nemotron}, Tsinghua University's ChatGLM~\cite{GLM-130B, du2021glm_du}, and TII's Falcon~\cite{penedo2023falcon} . New entries such as Command R ~\cite{cohere}, DBRX~\cite{DBRX}, Reka~\cite{ormazabal2024Reka}, Baichuan~\cite{Baichuan}, Qwen~\cite{Bai2023QwenTR}, Yi ~\cite{young2024yi}, DeepSeek~\cite{Bi2024DeepSeekLS}, InternLM~\cite{2023internlm}, MiniCPM~\cite{hu2024minicpm} and Llemma~\cite{azerbayev2023llemma} have also emerged. Presently, models like Alpaca~\cite{alpaca}, Vicuna~\cite{vicuna2023}, Guanaco~\cite{dettmers2023qlora}, Orca~\cite{mukherjee2023orca}, OpenChat~\cite{wang2023openchat}, Tulu2~\cite{ivison2023_tulu2}, WizardLM~\cite{xu2023wizardlm}, XwinLM~\cite{ni2024xwin, li2024xwinmath}, StarlingLM~\cite{zhu2023starling}  and Zephyr~\cite{Tunstall2023ZephyrDD} are being developed through supervised fine-tuning based on  Llama ~\cite{touvron2023llama,touvron2023llama2, Llama3}  and Mistral~\cite{jiang2023mistral}. However, how to measure the performance of current all models in real-world, open scenarios is a challenging task. LMSYS has developed a chatbot arena~\cite{chiang2024_lmsys_chatbot} that utilizes anonymous battle and human judgment, but assessing all models is both time-consuming and costly. In this paper we simulate an offline chatbot arena and employ advanced LLM (i.e., Llama3-70B-Chat~\cite{Llama3}) for judgment, significantly improving efficiency and reducing time requirements by 40x.

\subsection{\textbf{LLM Post-training}} 

The alignment performance of Large Language Models (LLMs) is significantly influenced by the quality of Supervised Fine-Tuning (SFT) data, which encompasses task difficulty~\cite{mukherjee2023orca}, query complexity~\cite{xu2023wizardlm, luo2023wizardcoder,luo2023wizardmath}, semantic diversity~\cite{ding2023_ultrachat,alpaca}, and sample size~\cite{zhou2024lima}. For instance, ~\cite{alpaca} generates diverse queries through self-instruct ~\cite{wang2022self_instruct} methods, while ~\cite{xu2023wizardlm, luo2023wizardcoder, luo2023wizardmath, zeng2024automatic} enhances model alignment by increasing query complexity. ~\cite{mukherjee2023orca} boosts NLP task performance by optimizing FLAN~\cite{longpre2023flan} queries and responses with specialized LLMs, and ~\cite{ding2023_ultrachat} has introduced UltraChat. To select data efficiently, some strategies like IFD~\cite{li2023_IFD}, INSTAG~\cite{lu2023instag}, DEITA~\cite{liu2023_deita}, MODS~\cite{du2023_MODS}, and ALPAGASUS~\cite{chen2023alpagasus} are adopted. ~\cite{mukherjee2023orca} employs ChatGPT to label instructional data, ensuring both diversity and complexity. Here, we select training data using the ``judge pair'' method with different advanced models.

To better adapt to preferences beyond SFT, models are trained with feedback-based methods like RLHF  and RLAIF~\cite{uesato2022deepmind-orms,lightman2023openai-verify-step-by-step,  ouyang2022training,bai2022constitutional-claude, touvron2023llama2}, employing Proximal Policy Optimization (PPO)~\cite{schulman2017_ppo} to align with model preferences. ~\cite{burns2023weak-to-strong-openai, ji2024aligner, chen2024SPIN} improve weak to strong model generalization. WizardMath~\cite{luo2023wizardmath} adopts RLEIF, introducing process supervision and instruction quality scoring reward model to improve the mathematical reasoning ability of large language models. Due to RLHF's complexity and instability, simpler alternatives like DPO~\cite{Rafailov2023_DPO}, RRHF~\cite{yuan2023rrhf}, KTO~\cite{Ethayarajh2024_KTO}, IPO~\cite{Azar2023_IPO}, sDPO~\cite{Kim2024_sDPO}, and ORPO~\cite{Hong2024_ORPO} are utilized. DPO~\cite{Rafailov2023_DPO} merges reward modeling with preference learning. RRHF~\cite{yuan2023rrhf} uses ranking loss to prioritize preferred answers, and KTO~\cite{Ethayarajh2024_KTO} operates without needing paired preference datasets. In this paper,  in order to efficiently manage massive data, we have established a dynamic data flywheel for model post-training through the pair-wise judge battle method to consistently collect feedback from the advanced models. Furthermore, we propose \methodname to perform iterative battle and training process (SFT-DPO-PPO), where the WizardLM-$\beta$ is continuously updated and re-evaluated against the SOTA models, progressively enhancing the performance of our model.

\subsection{\textbf{LLM Benchmarks}} 
Large Language Models (LLMs) have transformed the way people interact with computing systems and are extensively used in everyday life and work~\cite{zhao2023_llmsurvey}. The existing benchmarks~\cite{chang2024_llm_bench3_survey, guo2023_llm_bench2_evaluating, wang2024user_llmbench1} are mainly divided into two categories: 1) Specialized tasks. Knowledge and Capability: MMLU~\cite{hendrycks2020measuring}, CMMLU~\cite{li2023cmmlu}, and C-Eval~\cite{huang2024c}; Reasoning: ARC~\cite{clark2018think}, HellaSwag~\cite{zellers2019hellaswag},  PIQA~\cite{DBLP:conf/aaai/BiskZLGC20}, GSM8k~\cite{cobbe2021training_gsm8k_2}, MATH~\cite{hendrycks2021measuring}; Programming: HumanEval~\cite{chen2021evaluating}, MBPP~\cite{austin2021program}, LiveCodeBench~\cite{jain2024livecodebench}; Safety and Truthfulness: ToxicChat~\cite{lin2023toxicchat}, OLID~\cite{rosenthal2021solid}, BIG-Bench~\cite{srivastava2022_bigbench}, TruthfulQA~\cite{lin2021truthfulqa}. They focus on assessing LLM performance in specific areas.
2) General tasks: like MT-Bench~\cite{zheng2024judging_mtbench, bai2024mt} and AlpacaEval~\cite{alpaca_eval, dubois2024AlpacaEval2.0_LC, dubois2023alpacafarm}, encompass categories such as writing, role-playing, and mathematics, highlighting the models' comprehensive abilities and multi-turn dialogue performance.

Real-world benchmarks, (i.e.,  LMSYS ChatBot Arena~\cite{chiang2024_lmsys_chatbot} and Allenai WildBench~\cite{wildbench2024}) use anonymous battles, ELO~\cite{bai2022training,boubdir2023elo} rankings, and human judgments, but have time delay and often do not timely reflect the models' true performance and require large time and human labor intensive. ~\cite{li2024arena-hard-auto, zhao2024-Ali-auto-arena} propose an automatic evaluation tool for instruction-tuned LLMs. Additionally, most models overfit on leaderboards like MT-Bench~\cite{zheng2024judging_mtbench}, OpenLLM leaderboard~\cite{open-llm-leaderboard, eval-harness}, showing inconsistent performance with real-world ChatBot scenarios and low differentiation among models. Therefore, we have developed the simulated offline WizardArena, which not only effectively differentiates model performance but also aligns closely with the online human-based LMSYS 
 ChatBot Arena~\cite{chiang2024_lmsys_chatbot}, which achieves an average consistency of 98\% with LMSYS ChatBot
Arena, simultaneously making it suitable for selecting the optimal models and predicting the performance of models while significantly enhancing model post-training  through battle data.

\section{Conclusion}

This paper introduces \emph{Arena Learning}, a simulated offline chatbot arena that utilizes AI LLMs to bypass the manual and time-intensive cost typically associated with preparing the arena battle data, while preserving the core advantages of the arena-based evaluation and training. The effectiveness of \methodname is validated through the high consistency in predicting Elo rankings across various LLMs compared, when compared with the human-based LMSys Chatbot Arena. Furthermore, the model trained iteratively on synthetic data generated by \methodname exhibits significant performance improvements using various training strategies. 
Overall, \methodname emerges as a cost-effective and reliable alternative to conventional human-based evaluation systems, providing a sustainable approach to progressively enhance and scale the capabilities of large language models.

\textbf{Limitations and Broader Impacts.} If the judge model fails to accurately imitate human evaluators, the generated rankings and training data may be compromised. Moreover, similar to the other LLMs, our model could  generate potentially unethical or misleading information.

{
\small
\bibliography{neurips_2024}

}

\newpage
\appendix

\newpage
\section{Three consistency metrics between two Arenas}\label{appendix:consistency_metrics}

To more effectively align the online arena (i.e. LMSYS ChatBot Arena) with real-world human preferences and  to enhance differentiation among models, we developed a simulated offline arena.  This platform is designed to evaluate the actual performance of the models and to facilitate the selection of optimal model checkpoints.  We employ several key criteria~\cite{arenahard2024_lmsysarena_hard} that define an effective benchmark for evaluating Large Language Models (LLMs) in chatbot applications, aiming to enable meaningful functional comparisons across different models.

\begin{itemize}
  \item Alignment with Human Preference : The benchmarks should maintain high alignment with real-world human preferences in responses to the diverse and hard instructions, ensuring that the models' outputs meet user expectations.
  \item Ranking Accuracy: The benchmark should align closely with the reference standard to ensure that the rankings of different models on the leaderboard are reliable and accurate.
  \item Differentiation: The benchmark should be capable of accurately differentiating the performance of various models by providing confidence intervals with minimal overlap. This feature is crucial to ensure that the more effective models can be reliably distinguished.

\end{itemize}

We define the alignment of Benchmark \( A \) with reference to Benchmark \( B \), for a model pair \( (m_1, m_2) \) that \( B \) can confidently differentiate, using the following formulation:

The agreement score, \( s(m_1, m_2) \), is determined as:
\[
s(m_1, m_2) = 
\begin{cases} 
1.0 & \text{if } A \text{ confidently separates } m_1 \text{ from } m_2 \text{ and their ranking aligns with } B \\
-1.0 & \text{if } A \text{ confidently separates } m_1 \text{ from } m_2 \text{ and their ranking conflicts with } B \\
0.0 & \text{if } A \text{ cannot confidently separate } m_1 \text{ from } m_2
\end{cases}
\]

To assess ranking accuracy, we employed Spearman's rank correlation coefficient to analyze the correlation between the two sets of ranking data.

   \[
   \rho = 1 - \frac{6 \sum d_i^2}{n(n^2 - 1)}
   \]

where \( \rho \) is the Spearman's rank correlation coefficient, \( d_i \) is the difference between the ranks of corresponding variables, and \( n \) is the number of observations.

We define the differentiation of models based on their performance scores, which are represented by confidence intervals \( CI_1 \) and \( CI_2 \) via bootstrapping. If the two confidence intervals do not overlap,  then models \( M_1 \) and \( M_2 \) are considered to be separable.
\[CI_1 \cap CI_2 = \emptyset \]

\newpage

\end{document}